%% file: iclr2026_conference.tex
\documentclass{article} % For LaTeX2e
\usepackage{iclr2026_conference,times}

\input{math_commands.tex}

\usepackage{hyperref}
\usepackage{url}
\usepackage{graphicx}
\usepackage{booktabs}
\usepackage{subcaption}
\usepackage{wrapfig}
\usepackage{listings}
\usepackage{xcolor}
\usepackage{multirow}

\definecolor{codegray}{RGB}{245,245,245}

\lstdefinestyle{pythonstyle}{
  language=Python,
  basicstyle=\ttfamily\tiny,
  keywordstyle=\color{blue},
  stringstyle=\color{orange},
  commentstyle=\color{green!60!black},
  numbers=left,
  numberstyle=\tiny\color{gray},
  stepnumber=1,
  numbersep=5pt,
  showstringspaces=false,
  tabsize=4,
  breaklines=true,
  backgroundcolor=\color{codegray},
  frame=none,
  linewidth=0.95\linewidth,       % 设置代码宽度
  xleftmargin=0.05\linewidth,    % 左边空出 (1 - 0.95)/2 = 0.025
  xrightmargin=0.025\linewidth    % 右边同理
}

\lstdefinestyle{markdownstyle}{
  language=,
  basicstyle=\ttfamily\tiny,
  numbers=none,
  backgroundcolor=\color{codegray},
  frame=none,
  linewidth=0.95\linewidth,
  xleftmargin=0.025\linewidth,
  xrightmargin=0.025\linewidth,
  breaklines=true,           % 自动换行
  breakatwhitespace=true,    % 尽量在空格处换行
  postbreak=\mbox{},         % 去掉续行标记
  breakindent=0pt            % 换行不缩进
}

\title{VisionLaw: Inferring Interpretable Intrinsic Dynamics from Visual Observations via Bilevel Optimization}

\author{Jiajing Lin$^{1}$, Shu Jiang$^{2}$, Qingyuan Zeng$^{2,3}$, Zhenzhong Wang$^{1}$, Min Jiang$^{1}$\thanks{Corresponding author: Min Jiang, minjiang@xmu.edu.cn} \\
$^{1}$School of Informatics, Xiamen University. \\
$^{2}$Institute of Artificial Intelligence, Xiamen University. \\
$^{3}$The Hong Kong University of Science and Technology (Guangzhou) \\
\texttt{\{jiajinglin, shujiang, 36920221153145\}@stu.xmu.edu.cn} \\
\texttt{\{zhenzhongwang, minjiang\}@xmu.edu.cn}
}

\iclrfinalcopy
\begin{document}

\maketitle

\begin{abstract}
The intrinsic dynamics of an object governs its physical behavior in the real world, playing a critical role in enabling physically plausible interactive simulation with 3D assets. 
% Existing methods have attempted to infer the intrinsic dynamics of objects from visual observations, but generally face two major challenges: one line of work relies on manually defined constitutive priors, making it difficult to generalize to complex scenarios; the other models intrinsic dynamics using neural networks, resulting in limited interpretability and poor generalization. 
Existing methods have attempted to infer the intrinsic dynamics of objects from visual observations, but generally face two major challenges: one line of work relies on manually defined constitutive priors, making it difficult to align with actual intrinsic dynamics; the other models intrinsic dynamics using neural networks, resulting in limited interpretability and poor generalization.
To address these challenges, we propose VisionLaw, a bilevel optimization framework that infers interpretable expressions of intrinsic dynamics from visual observations.
At the upper level, we introduce an LLMs-driven decoupled constitutive evolution strategy, where LLMs are prompted to act as physics experts to generate and revise constitutive laws, with a built-in decoupling mechanism that substantially reduces the search complexity of LLMs.
At the lower level, we introduce a vision-guided constitutive evaluation mechanism, which utilizes visual simulation to evaluate the consistency between the generated constitutive law and the underlying intrinsic dynamics, thereby guiding the upper-level evolution.
Experiments on both synthetic and real-world datasets demonstrate that VisionLaw can effectively infer interpretable intrinsic dynamics from visual observations. It significantly outperforms existing state-of-the-art methods and exhibits strong generalization for interactive simulation in novel scenarios. Our implementation is available at \href{https://github.com/JiajingLin/VisionLaw}{\texttt{github.com/JiajingLin/VisionLaw}}.
\end{abstract}

\vspace{-10pt}

\section{Introduction}
With the advancement of 4D generation~\cite{Animate124,4D-FY,Consistent4D,DreamGaussian4D}, realistic interaction with 3D assets has become increasingly feasible, facilitating broad applications in areas like virtual reality, embodied intelligence, and animation~\cite{VR-GS, EA1, EA2, sun2024conceptfactory, sun2025physically}. Among these advances~\cite{Physgaussian,Phys4DGen}, incorporating physical simulation~\cite{MPM, PBD} stands out as a particularly prominent method, as it enables the generation of interactive dynamics that closely mirror real-world physical behavior. To ensure simulation realism, it is essential to accurately capture the intrinsic dynamics of objects, including material properties (e.g., stiffness) and constitutive laws~\cite{continuum}, which describe the response behaviors of materials under applied forces.

Humans can roughly infer the intrinsic dynamics of objects merely by observing their motion, and are even capable of predicting how these objects would interact in new scenarios. A fundamental question arises: {\em can we enable machines to infer the intrinsic dynamics directly from visual observations, as humans do?} Recent methods~\cite{Physgaussian, PAC-NeRF} have attempted to bridge the gap between visual dynamics and physical simulation by incorporating physical simulators (e.g., Material Point Method, MPM~\cite{MPM}) into 3D representations such as NeRF and 3D Gaussian Splatting (3DGS)~\cite{NeRF,3DGS}. 
This integration has led to a promising paradigm for inferring the intrinsic dynamics from visual observations.
% typically by optimizing these properties embedded in a simulator so that the simulated visual dynamics from a differentiable renderer closely match the observed ones.
% 润色语言
Depending on the type of intrinsic dynamics being inferred, existing methods can be categorized into two groups: material parameter estimation and constitutive law inference.

% \begin{wrapfigure}{r}{0.5\linewidth} % r 表示右侧环绕，l 表示左侧
%   \centering
%   \vspace{-10pt} % 可选：调整图像与上文的垂直间距
%   \includegraphics[width=1.0\linewidth]{image/intro_new_crop.pdf}
%   \caption{{\em VisionLaw} infers interpretable intrinsic dynamics from visual observations via a bilevel optimization framework. At the upper level, we prompt LLMs as a thinker to generate constitutive hypotheses. At the lower level, we validate these hypotheses through visual simulation and generate feedback to guide hypothesis evolution
%   at the upper level.}
%   \vspace{-10pt} % 可选：调整图像与下文的垂直间距
% \end{wrapfigure}

\begin{figure}[!t]
	\centering
	\includegraphics[width=1\linewidth]{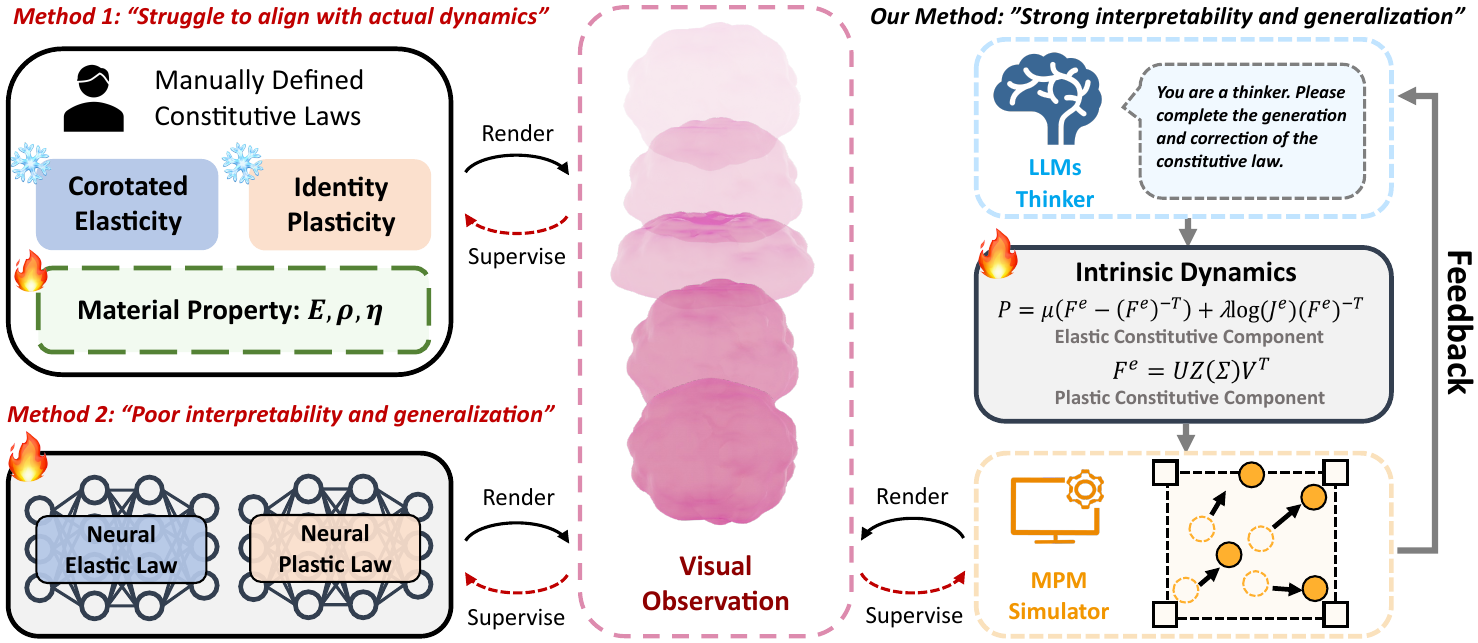}
	\caption{Existing works either rely on manually defined constitutive laws, which struggle to align with actual intrinsic dynamics, or learn neural constitutive laws, which suffer from poor interpretability and generalization. In contrast, our approach can automatically infer interpretable intrinsic dynamics solely from visual observations.}
    \vspace{-10pt}
\end{figure}

For material parameter estimation, PAC-NeRF and GIC~\cite{PAC-NeRF,GIC} estimate material parameters by the supervision of multi-view videos. PhyDreamer, DreamPhysics, and Physics3D~\cite{PhysDreamer,DreamPhysics,Physics3D} distill visual dynamics priors from video diffusion models to guide the estimation process. 
% To address the limitations of color loss in scenarios involving large motion, PhysFlow~\cite{PhysFlow} introduces optical flow loss. 
However, these approaches typically rely on manually defined constitutive laws, which often fail to align with the complex physical behaviors observed in practice, thereby compromising the accuracy of parameter estimation.

For constitutive law inference,
% OmniPhysGS~\cite{OmniPhysGS} introduces Constitutive Gaussians, which leverage temporal score distillation sampling (T-SDS)~\cite{DreamFusion} to assign an suitable constitutive model for each Gaussian kernel from a predefined model set. However, this expert-designed constitutive model set often fails to cover the full spectrum of possible dynamic behaviors. 
OmniPhysGS~\cite{OmniPhysGS} introduces constitutive Gaussians, which assign a suitable constitutive law to each Gaussian kernel from an expert-designed constitutive set. However, such a predefined set often fails to capture the full diversity of real-world physical behaviors.
% NeuMA~\cite{NeuMA} learns neural constitutive models to align with visual observations, and while it has shown promising results, several critical challenges remain: The neural network as a black box, lacking interpretability and making the constitutive model difficult for humans to understand. 
% NeuMA~\cite{NeuMA} learns neural constitutive laws to align with visual observations. Despite their effectiveness, these laws are represented as a black-box formulation, lacking interpretability, and thus making them difficult for humans to understand.
NeuMA~\cite{NeuMA} learns neural constitutive laws from visual observations. Despite its effectiveness, it has notable limitations: 
1) The learned laws are black-box representations, which lack interpretability and are difficult for humans and LLMs to understand; 
% 2) due to the sparsity of supervision information in visual observations and the absence of explicit physical inductive biases, neural networks are prone to overfitting, resulting in limited generalization.
% 2) due to the lack of explicit physical inductive biases, neural networks are prone to overfitting, resulting in limited generalization.
2) Due to the lack of physical inductive biases, neural networks tend to mechanically reconstruct visual observations instead of modeling the underlying dynamics, resulting in overfitting and poor generalization.
% This limitation makes them prone to overfitting and thus poor in generalization.

% 润色语言
% 2) 
% % due to information sparsity of visual observations, 
% The model is prone to overfitting, which hinders its ability to capture underlying physical laws and leads to poor generalization when applied to novel scenarios for interaction generation.

To overcome the aforementioned challenges, we introduce {\em VisionLaw}, an interpretable intrinsic dynamics inference framework based on bilevel optimization, which can jointly infer symbolic constitutive laws and their corresponding continuous material properties solely from visual observations.
At the upper level, we propose an LLMs-driven decoupled constitutive evolution strategy, which: 
1) unleashes the capabilities of LLMs in physical understanding and mathematical reasoning to generate and refine symbolic constitutive hypotheses; 
% 1) unleashes the capabilities of LLMs in both introducing physical inductive biases and mathematical reasoning, to generate and refine constitutive hypotheses; 
2) introduces a decoupling mechanism to effectively alleviate the search space explosion caused by jointly evolving elastic and plastic components.
% thereby improving both search efficiency and solution convergence.
% At the lower level, we construct a Vision-Guided Constitutive Evaluation mechanism, that uses differentiable simulation and rendering to optimize material parameters under a given constitutive law, aiming to reproduce the intrinsic dynamics underlying visual observations. The results reflect the intrinsic dynamics consistency of the constitutive law and are fed back to the upper level.
% At the lower level, we construct a Vision-Guided Constitutive Evaluation mechanism. Under visual observation supervision, the continuous material parameters of a given constitutive law are optimized via a differentiable simulator and renderer, with the goal of identifying the optimal configuration and generating evaluation and feedback that reflects the intrinsic dynamics consistency of the law, which in turn guides the upper-level evolution.
At the lower level, we construct a vision-guided constitutive evaluation mechanism. Supervised by visual observations, it optimizes the continuous material parameters of a given constitutive law using a differentiable simulator and renderer. The goal is to generate evaluation and feedback that reflect the consistency between the generated laws and ground-truth intrinsic dynamics, which in turn guides the evolution at the upper level.
% 润色
% Through collaborative optimization between the upper and lower levels, {\em VisionLaw} effectively captures the interpretable intrinsic dynamics underlying visual observations, which can be effectively applied to generate 4D interaction in novel scenarios. Our contributions are summarized as follows:
% Through collaborative optimization between the upper and lower levels, {\em VisionLaw} effectively captures the interpretable intrinsic dynamics from visual observations and generalizes them to novel scenarios, enabling physically plausible 4D interaction. 
% Our contributions are summarized as follows:
% \begin{itemize}
% \item We propose a bilevel optimization framework that can automatically infer symbolic constitutive law and material properties from visual observations.
% \item We distill physics priors from LLMs to introduce explicit physical inductive bias, thereby facilitating the evolution of constitutive laws. In addition, a decoupled evolution strategy is introduced to significantly improve both search efficiency and solution quality.
% % 利用大语言模型理解、处理
% \item We introduce a vision-guided constitutive evaluation mechanism to provide evaluation and feedback of a given constitutive law for the upper-level evolution.
% % that captures the intrinsic dynamic consistency of the law.
% \item Extensive experiments on both synthetic and real-world datasets demonstrate that our method effectively captures the interpretable intrinsic dynamics underlying visual observations and transfers them to novel scenarios for 4D interaction.
% \end{itemize}
% \textblue{
Our contributions are summarized as follows:
\begin{itemize}
    \item We propose a bilevel optimization framework that unifies constitutive evolution and vision-guided constitutive evaluation, achieving the inference of symbolic constitutive laws and material properties from visual observations.
    \item We introduce physical inductive biases through LLMs to guide the evolution of constitutive laws. In addition, a decoupled evolution strategy is proposed to markedly improve both search efficiency and solution quality.
\end{itemize}
Extensive experiments on both synthetic and real-world datasets demonstrate that our method effectively captures the interpretable intrinsic dynamics underlying visual observations and generalizes them to novel scenarios for 4D interaction.

\section{Preliminaries}
\subsection{Constitutive laws and Material Point Method}
In continuum mechanics~\cite{continuum}, constitutive laws define how materials respond under applied forces.
The essential reason why materials like rubber, sand, and water exhibit entirely different physical behaviors lies in the differences in the constitutive laws they follow.
To simulate the motion and deformation of materials, we need to solve a system of partial differential equations derived from the conservation of mass and momentum:
\begin{equation}
\frac{D\rho}{Dt} + \rho \nabla \cdot \mathbf{v} = 0, \quad
\rho \frac{D\mathbf{v}}{Dt} = \nabla \cdot \mathbf{P} + \rho \mathbf{g},
\end{equation}
where $\rho$ denotes the density, $\mathbf{v}$ the velocity field, $\mathbf{g}$ the gravitational acceleration, and $\mathbf{P}$ the stress tensor, which is defined by the constitutive law.
% \textblue{
The system becomes closed and solvable only after a specific constitutive relation for $\mathbf{P}$ is prescribed.
% }

% In this paper, we employ the MPM simulator to solve the above system of governing equations for simulation. 
% % 
% Please refer to Appendix~\ref{appendix: MPM} for further details about MPM. 

% \textblue{
In this paper, we employ an MPM simulator~\cite{MPM} to solve the above governing equations because of its versatility in handling various materials. Intuitively, the MPM discretizes the continuum into a set of material points that carry physical quantities such as mass, velocity, and deformation gradient, and updates system states through particle–grid transfers and time integration. At each time step, the particles first project their quantities onto a background grid; the momentum conservation equation is then solved on the grid to compute nodal velocities and accelerations; finally, these updated grid quantities are interpolated back to the particles, advancing their positions $\mathbf{x}$ and deformation gradients $\mathbf{F}$, which describes the local deformation. For more details about MPM, please refer to Appendix~\ref{appendix: MPM}.
% }

Within the MPM framework, two types of constitutive laws must be specified: (1) an elastic constitutive law that describes reversible elastic responses, and (2) a plastic constitutive law that captures irreversible plastic evolution. Their formulations are given as:
\begin{equation}
\varphi_E\left(\mathbf{F}; \theta_E\right) \mapsto \boldsymbol{\tau}, \quad
\varphi_P\left(\mathbf{F}; {\theta}_P\right) \mapsto \mathbf{F}^{\text{corrected}},
\end{equation}
where $\varphi_E$ and $\varphi_P$ denote the elastic and plastic constitutive laws, respectively. $\mathbf{F}$ is the deformation gradient, $\boldsymbol{\tau}$ is the Kirchhoff stress tensor, ${\mathbf{F}^{\text{corrected}}}$ is the corrected deformation gradient after plastic return mapping. The continuous material parameters in the elastic and plastic laws are denoted by $\theta_E$ and $\theta_P$, respectively. Several classical constitutive laws are listed in Appendix~\ref{appendix: expert-designed constitutive laws}. 
Despite the availability of many classical constitutive laws, they remain inadequate in capturing the diversity and nonlinear behavior of complex materials. To this end, we propose {\em VisionLaw}, which infers constitutive laws directly from visual observations. 
% This enables accurate modeling of the intrinsic dynamics and facilitates the interactive simulation in novel scenarios.

\subsection{Physics-Integrated 3D Gaussians}
\label{Physics-Integrated 3D Gaussians}
3D Gaussians Splatting (3DGS)~\cite{3DGS} represents the scene using a set of anisotropic Gaussian kernels $\mathcal{G}=\{\mathbf{x}_i, \mathbf{A}_i, \alpha_i, \mathbf{\mathcal{C}}_i\}_{i \in \mathcal{K}}$, where $\mathbf{x}_i$, $\mathbf{A}_i$, $\alpha_i$, and $\mathbf{\mathcal{C}}_i$ represent the center position, covariance matrix, opacity, and spherical harmonic coefficients of the Gaussian kernel $\mathcal{G}_i$, respectively. To render 3D Gaussians into a 2D image from a given view, the color of each pixel can be formulated as:
\begin{equation}
\label{eq:render}
    \mathbf{C} = \sum_{i \in \mathcal{N}}\sigma_i\mathbf{SH}(d_i,\mathcal{C}_i)\prod_{j=1}^{i-1}(1-\sigma_j),
\end{equation}
where $\mathcal{N}$ denotes a set of sorted Gaussian kernels related to the pixel and view. $\sigma_i$ is the effective opacity, defined as the product of the projected 2D Gaussian weight and opacity $\alpha_i$. $\mathbf{SH}$ computes RGB values based on the view direction $d_i$ and spherical harmonic coefficients $\mathbf{\mathcal{C}}_i$.
% Unlike NeRF's implicit form, 3DGS offers an explicit representation that not only significantly accelerates training and rendering, but also inherently exhibits a Lagrangian nature, facilitating seamless integration with simulation algorithms. 
Unlike NeRF's implicit form, 3DGS offers an explicit representation that exhibits a Lagrangian nature, facilitating seamless integration with simulation algorithms.
Thus, PhyGaussians~\cite{Physgaussian} pioneers the integration of MPM simulator~\cite{MPM} into 3DGS, combining physical simulation with visual rendering.
Specifically, this method treats Gaussian kernels as particles representing the continuum, and assigns each a time property $t$, material properties $\theta$ (e.g., stiffness).
% \begin{equation}
%     \mathcal{G}^t=\{\mathbf{x}_i^t, \mathbf{A}_i^t, \alpha_i, \mathbf{\mathcal{C}}_i,\theta_i\}_{i \in \mathcal{K}}.
% \end{equation}
Therefore, given the constitutive law and simulation conditions (e.g., external forces and boundary), MPM can be applied to predict the displacement and deformation of Gaussian kernels at the next time step:
\begin{align}
    \mathbf{x}^{t+1}, \mathbf{F}^{t+1} &= \mathbf{\Phi}(\mathcal{G}^t),
    \\
    \mathbf{A}^{t+1} = \mathbf{F}&^{t+1}\mathbf{A}^t(\mathbf{F}^{t+1})^T.
\end{align}
% \textblue{
Here, $\mathbf{\Phi}$ is a differentiable MPM simulator, $\mathbf{F}^{t+1}$ denotes the deformation gradient at time step $t+1$ 
% which describes the local deformation of particles 
(the subscript $i$ is omitted for simplicity).
Gaussian covariance $\mathbf{A}^{t+1}$ can be updated by applying $\mathbf{F}^{t+1}$, which approximates the deformation of the Gaussian kernel. After completing the MPM simulation, a 4DGS representation is constructed, which enables rendering of visual dynamics using Eq.~\ref{eq:render}. 
% }
% The rendered dynamics can be further compared against visual observations to compute supervised loss, enabling back-propagation to optimize learnable factors within the MPM system.

\begin{figure}[!t]
\centering
\includegraphics[width=1\linewidth]{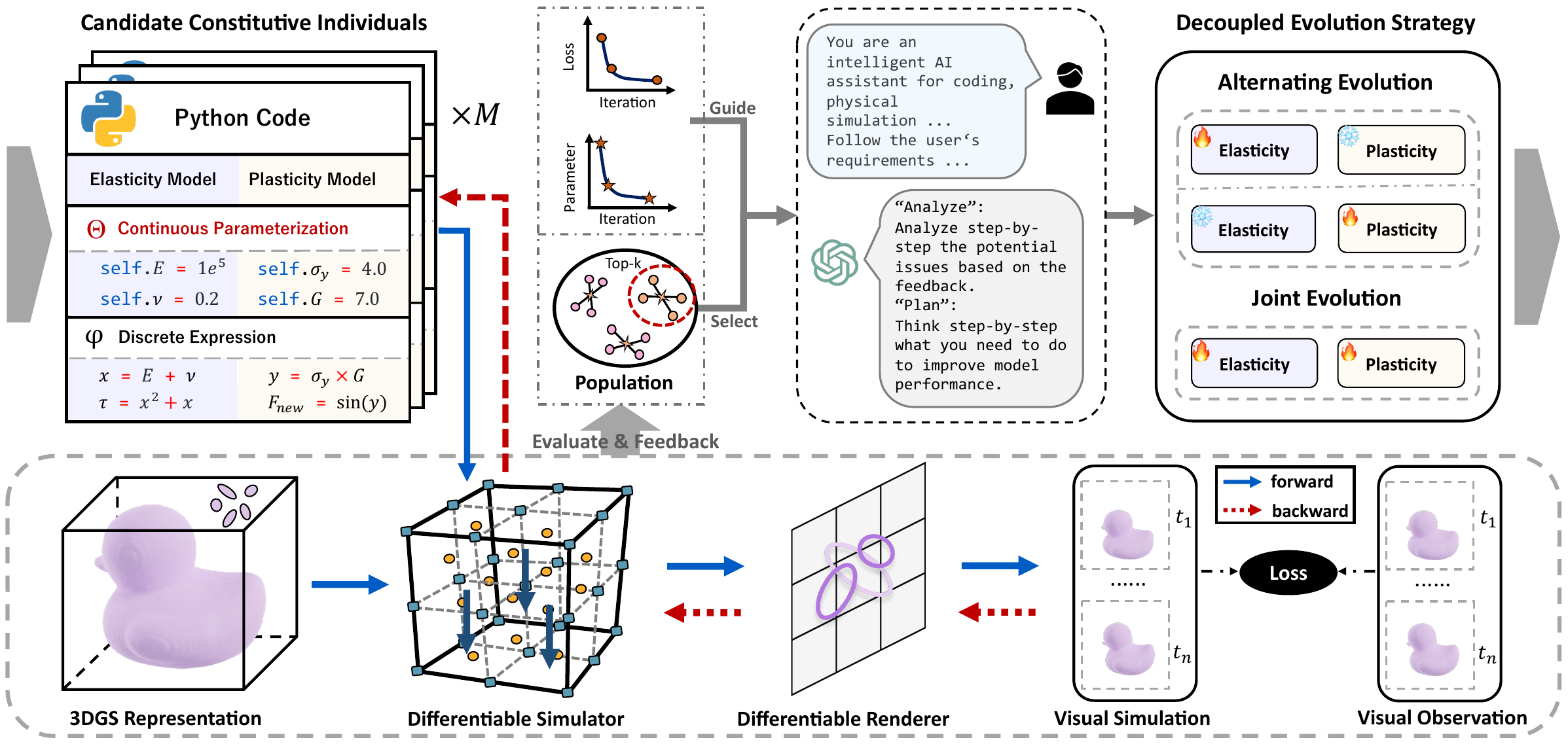}
\caption{
Given a constitutive individual—either predefined at initialization or generated by LLMs—it is embedded into a differentiable MPM simulator for forward simulation. The resulting dynamics are rendered and compared with observations to compute a loss, which is backpropagated to optimize material parameters. This process produces both a fitness score and feedback for the individual. Based on fitness, the top-k individuals are selected and, along with their feedback, encoded into prompts for the LLMs. Guided by the decoupled evolution strategy, the LLMs analyze and refine these constitutive law expressions to generate offspring for the next optimization cycle.
}
% \vspace{-5pt}
\label{pipeline}
\end{figure}

\section{Methodology}
% \subsection{Problem Definition and Overview}
In this work, we aim to infer interpretable intrinsic dynamics from a series of visual observations.
Formally, given multi-view video observations $V = \{V_1, V_2, ..., V_N\}$ of moving objects along with corresponding camera extrinsic and intrinsic parameters, the goal is to infer the discrete constitutive law expressions and optimize the continuous material parameters in a unified manner. 
% However, discrete structure search and continuous parameter optimization are fundamentally different in nature, making it difficult to unify them within a single end-to-end framework. 
To this end, we propose {\em VisionLaw}, a novel bilevel optimization framework:
% \begin{equation}
%     \min_{\varphi_E,\varphi_P,\theta_E^*,\theta_P^*} \; \mathcal{L} \left( \mathcal{R}\left(\varphi_E,\varphi_P,\theta_E^*,\theta_P^*; \Phi,\mathcal{G}\right), V\right)
% \end{equation}
% \begin{equation}
%     \text{s.t.} \quad \psi \left( \varphi_E,\varphi_P,\theta_E,\theta_P; \Phi \right) \leq 0
% \end{equation}
% \begin{equation}
%     \theta^* \in \arg\min_{\theta \in \Theta} \mathcal{L} \left( \mathcal{R}\left( \theta_E,\theta_P; \Phi, \varphi_E,\varphi_P \right) \right)
% \end{equation}
\begin{equation}
    \min_{\varphi,\Theta} \; \mathcal{L} \left( \mathcal{R}\left(\varphi,\Theta,\theta^*; \Phi,\mathcal{G}\right), V\right),
\end{equation}
\begin{equation}
    \text{s.t.} \quad h \left( \varphi,\Theta; \Phi \right) \leq 0,
\end{equation}
\begin{equation}
\label{eq:parameter optimization}
    \theta^* \in \arg\min_{\theta \in \Theta} \mathcal{L} \left( \mathcal{R}\left( \theta; \varphi, \Phi, \mathcal{G} \right), V \right),
\end{equation}
where $\mathcal{R}$ is a differentiable renderer defined by Eq.~\ref{eq:render}. 
The constitutive law $\varphi$ consists of an elastic law $\varphi_E$ and a plastic law $\varphi_P$.
$\Theta$ defines the continuous parameter space for inner-level optimization $\theta \in \Theta$.
$h(\cdot) \leq 0$ refers to the validity of the simulation (e.g. whether a constitutive law $\varphi$ is simulatable). The material parameter $\theta$ includes the elastic parameters $\theta_E$ and the plastic parameters $\theta_P$.
% and $\theta^*$ denotes the optimal continuous parameters.
For the upper level, based on evaluation and feedback from the lower level, LLMs are employed to generate and refine constitutive law expressions $(\varphi,\Theta)$.
At the lower level, given the output $(\varphi,\Theta)$ from the upper level, the optimal continuous material parameters $\theta^*$ are estimated under visual observation supervision, using differentiable rendering and simulation. During this process, evaluation and feedback are provided.
% to guide the upper-level evolution.
% The pipeline of the proposed xxx is illustrated in Fig.~\ref{xxx}, and the pseudocode of the method is provided in Algorithm~\ref{xxx}.
The pipeline of the proposed {\em VisionLaw} is illustrated in Fig.~\ref{pipeline}.
% and the corresponding pseudocode is presented in Algorithm\ref{xxx}.

% \subsection{LLMs-Driven Decoupled Constitutive Evolution}
\subsection{Upper-Level Constitutive Laws Evolution}
\label{sec:upper level}
\subsubsection{LLMs-Driven Constitutive Evolution}
Recently, LLMs have shown tremendous potential in scientific discovery~\cite{llms1,llms2,SGA, wzz-1, wzz-2}, owing to their strong symbolic reasoning abilities and extensive physical priors. 
% \textblue{
Inspired by this, in the upper-level search, we prompt LLMs to evolve discrete constitutive law expressions. 
% For details on prompt design, please refer to Appendix xxx.
Specifically, we consider LLMs as an intelligent operator and construct an evolutionary search paradigm to iteratively optimize the constitutive law expressions. 
% }
Each law is represented as a Python code snippet with a clear physical meaning and strong interpretability.

The optimization procedure consists of five stages, which are as follows:
% \textblue{
i) Initialization: Classical constitutive laws (e.g., purely elastic material models) are introduced as initial individuals. This serves as a physically plausible starting point for the evolutionary process.
% }
% \textblue{
ii) Fitness Evaluation: Each candidate constitutive individual is passed to the lower level for simulation testing, 
% Its fitness is evaluated based on visual observation, and feedback, such as the loss curve, is collected.
where its fitness is evaluated from visual observations and the corresponding feedback information is collected.
% }
iii) Selection: To improve population diversity and avoid local optima, we first remove duplicate constitutive individuals with fitness differences below a threshold $\epsilon$. Then, we select the top-k constitutive individuals with the highest fitness from the remaining population as "parents" for the next round of evolution.
iv) Expression Correction: We prompt LLMs to 1) analyze the parent expression and identify any shortcomings based on its feedback; 2) design an improvement plan and determine how to modify the expression to increase fitness; 3) generate a set of physically plausible constitutive law expressions as candidate individuals.
% which are incorporated into the population as "offspring." 
This process is formalized as:
\begin{equation}
\{\varphi^m, \Theta^m\}_{m \in |M|} = \text{LLM} \left( \left\{ \mathcal{\varphi}^k, \Theta^k, \mathcal{O}^k  \right\}_{k \in |K|}, \mathcal{P}\right),
\end{equation}
where $K$ denotes parent size, $M$ denotes offspring size, $\mathcal{O}$ represents the feedback obtained from the lower level and $\mathcal{P}$ denotes the prompt provided to LLMs.
v) Iteration: Repeat steps ii) to iv) until a predefined number of iterations is reached.
% v) Stages ii) to iv) constitute a complete evolutionary iteration. Multiple evolutionary iterations are executed until the predefined number of iterations is reached.
Eventually, the algorithm discovers constitutive laws that not only simulate dynamic behaviors consistent with visual observations but also exhibit strong physical interpretability.

\subsubsection{Decoupled Evolution Strategy}
In the MPM simulation framework, a complete constitutive law $\varphi$ consists of an elastic part $\varphi_E$ and a plastic part $\varphi_P$, which together govern the system’s simulation behavior.
However, simultaneous optimization of these components significantly enlarges the search space, increases the difficulty of LLMs search, and hinders convergence to high-quality solutions.
To address the above issue, we propose a decoupled evolution strategy that splits the coupled constitutive optimization task into two independently solvable sub-tasks, thereby effectively reducing the search space. 

This strategy consists of two phases: 1) Alternating Evolution: In each iteration, we prompt the LLM to optimize only one component of the constitutive law expression (elastic or plastic), 
% while keeping the other fixed to be updated in the next iteration. 
while the other remains fixed and is updated in the subsequent iteration.
% The two types of constitutive law expressions are optimized in an alternating manner over multiple iterations. 
The two components of constitutive laws are optimized alternately across multiple iterations.
% \begin{equation}
% \{\varphi_{E}^m, \Theta_{E}^m\}_{m \in |M_E|} = \text{LLM} \left( \left\{ \mathcal{\varphi}^k, \Theta^k, o^k  \right\}_{k \in [K_E]}\right),
% \end{equation}
% \begin{equation}
% \{\varphi_{P}^m, \Theta_{P}^m\}_{m \in |M_P|} = \text{LLM} \left( \left\{ \mathcal{\varphi}^k, \Theta^k, o^k  \right\}_{k \in [K_P]} \right).
% \end{equation}
2) Joint Evolution: After the alternating optimization phase, we prompt the LLM to jointly optimize both elastic and plastic components to further enhance overall performance. This phase serves as a fine-grained refinement of the existing high-quality expressions from a global perspective. Through the proposed decoupled evolution strategy, we effectively reduce the search space, enhance the stability and efficiency of LLM-based search, and substantially improve the quality of the final constitutive laws.

% \subsection{Vision-Guided Constitutive Evaluation}
\subsection{Lower-Level Constitutive Laws Evaluation}
\label{sec:lower level}
To effectively evaluate whether a candidate constitutive expression can accurately capture the intrinsic dynamics of motion observed in visual data and to provide high-quality feedback to the upper-level evolution, we design a vision-guided constitutive evaluation mechanism.
% we design a vision-supervised continuous material parameter optimization framework.
First, a static 3DGS representation is reconstructed from the first frame of multi-view video inputs. 
Then, the candidate constitutive law expression \( \varphi(\theta)\), with continuous material parameters, is seamlessly embedded into a differentiable MPM simulator. 
% Following the paradigm described in~\ref{Physics-Integrated 3D Gaussians}, 
We integrate the MPM simulator with 3DGS to drive the simulation and render the predicted visual dynamics \( V \) from given views. The supervised loss between the predicted and observed visual dynamics can be formulated as:
\begin{equation}
\mathcal{L} = \frac{1}{N}\sum_{n=1}^{N}[\lambda \mathcal{L}_2(\hat{V}_n, V_n) + (1 - \lambda) \mathcal{L}_{\text{D-SSIM}}(\hat{V}_n, V_n)],
\end{equation}
where $\hat{V}_n$ denotes the rendered video from the $n$-th viewpoint, and $\mathcal{L}_2$ is the L2 norm loss.
Since both the renderer $\mathcal{R}$ and the MPM simulator $\Phi$ are differentiable, the evaluation loss can be backpropagated to optimize the continuous material parameters $\theta$ as described in Eq.~\ref{eq:parameter optimization}. During this process, we collect the loss curve and the material parameter update trajectory as feedback $\mathcal{O}$ to construct the LLMs' prompts. Meanwhile, the minimum loss achieved during optimization is used as the fitness score of the constitutive candidate to guide the selection process at the upper level.

% MPM simulator can be regarded as a sequential model similar to RNN~\cite{RNN}, which is prone to gradient vanishing or explosion in optimization.
% To mitigate this issue, we adopt a Truncated Backpropagation Through Time (BPTT) strategy, where gradient propagation is limited to a fixed number of consecutive frames, thereby improving training stability.
% Specifically, given a video sequence containing $N$ frames, we divide it into $T$ consecutive temporal segments $(T>1)$, each containing $\frac{N}{T}$ frames used for supervision and gradient backpropagation.
% In each optimization iteration, we perform $T$ steps of local backpropagation and parameter updates. This strategy effectively mitigates the training instability caused by the long-horizon nature of MPM simulation, leading to a more stable and controllable parameter optimization.

\begin{table}[t]
\centering
\resizebox{\linewidth}{!}{%
\begin{tabular}{l|cccccc|c}
\toprule
\textbf{Method} & \textbf{BouncyBall} & \textbf{ClayCat} & \textbf{HoneyBottle} & \textbf{JellyDuck} & \textbf{RubberPawn} & \textbf{SandFish} & \textbf{Average} \\
\midrule
PAC-NeRF~\cite{PAC-NeRF} & 516.30 & 15.38 & 2.21 & 137.73 & 15.47 & 1.71 & 114.80 \\
NCLaw~\cite{NCLaw} & 56.69 & 2.35 & 0.92 & 11.97 & 3.91 & 1.30 & 12.86 \\
NeuMA~\cite{NeuMA} & 1.78 & 1.24 & 1.09 & 10.96 & 1.01 & \textbf{1.07} & 2.86 \\
\midrule
VisionLaw (Ours) & \textbf{1.08} & \textbf{0.77} & \textbf{0.79} & \textbf{5.19} & \textbf{0.94} & {1.10} & \textbf{1.65} \\
\bottomrule
\end{tabular}
}
\caption{\textbf{Quantitative Comparison of Intrinsic Dynamics Consistency on Synthetic Datasets.} The Chamfer distance was employed to quantify the similarity between simulated and ground-truth particle trajectories. Lower values indicate better alignment with ground-truth intrinsic dynamics.}
\label{tab:chamfer distance}
\vspace{-10pt}
\end{table}

\section{Experiments}
\subsection{Experimental Setup}
\subsubsection{Implementation Details}
Given multi-view videos of a scene, we follow NeuMA~\cite{NeuMA} to perform 3D reconstruction and Particle-GS binding using multi-view images from the initial time step.
We use only single-view videos as ground-truth observations to infer intrinsic dynamics across all experiments.
For all scenarios, the initial constitutive individual is only defined as a purely elastic model that combines linear isotropic elasticity with identity plasticity.
For the upper-level evolution, we employ \texttt{GPT-4.1-mini} to generate constitutive hypotheses. 
Details of the prompt design are provided in Appendix~\ref{appendix: prompt design details}.
The decoupled evolution strategy is executed through four iterations of alternating optimization, followed by three iterations of joint optimization.
For lower-level optimization, we conduct MPM simulation~\cite{Physgaussian} under gravitational acceleration ($9.8 m/s^2$). 
We employ the Adam optimizer with a learning rate of $1 \times 10^{-3}$ to tune the material parameters.
For each scene, we perform five independent runs using different random seeds.
All experiments are conducted on an NVIDIA A40 (48 GB) GPU. Detailed experimental settings are provided in Appendix~\ref{appendix: implementation details}.

\subsubsection{Baselines}
% We evaluated our proposed method by comparing it with state-of-the-art intrinsic dynamics inference methods: PAC-NeRF~\cite{PAC-NeRF}, NCLaw~\cite{NCLaw}, NeuMA~\cite{NeuMA}, and Spring-Gaus~\cite{Spring-Gaus}. 
We compare our method with state-of-the-art intrinsic dynamics inference methods: PAC-NeRF~\cite{PAC-NeRF}, NCLaw~\cite{NCLaw}, NeuMA~\cite{NeuMA}, and Spring-Gaus~\cite{Spring-Gaus}.
PAC-NeRF is capable of inverting material parameters from video input.
NCLaw only fits neural constitutive laws to known dynamics, whereas NeuMA extends this by introducing visual information for adaptation.
NeuMA is the most relevant work to ours, as it learns neural constitutive laws directly from visual observations.
% 此处后半句为新添加的
Spring-Gaus models elastic objects by integrating spring-mass system with Gaussian kernels. It achieves intrinsic dynamics inference through the estimation of spring stiffness.
All baseline experimental settings follow the original setup.
% We compare it with our method in real-world data.

% We adopt it as a baseline across all datasets.

\subsubsection{Datasets and Metrics}
To thoroughly evaluate the effectiveness of our method, we conduct experiments on both synthetic and real-world datasets. 
For synthetic data, we adopt six dynamic scenes from NeuMA~\cite{NeuMA}, each with varying initial conditions (including object shapes, velocities, and positions), intrinsic dynamics, and simulation time intervals. 
% 后面新增的内容
% The scenarios are denoted as: 'BouncyBall', 'JellyDuck', 'RubberPawn', 'ClayCat', 'HoneyBottle', and 'SandFish'. 
Each synthetic scene includes 10 videos captured from different views, each containing 400 frames, and the dataset further provides ground-truth particle trajectories.
For real-world evaluation, we conduct experiments on two scenes ('Bun' and 'Burger') provided by Spring-Gaus~\cite{Spring-Gaus}. 
Each real-world scene includes 3 videos captured from different views, each containing 19 frames.
More details of the datasets are provided in Appendix~\ref{appendix: dataset details}.
Following prior works~\cite{Neurofluid,NeuMA}, we use the L2-Chamfer distance~\cite{Chamfer} between the simulated and ground-truth particle trajectories to quantify the accuracy of intrinsic dynamics inference. To assess the visual fidelity, we follow 3DGS~\cite{3DGS} and employ PSNR, SSIM, and LPIPS as quantitative metrics. 

\begin{figure}[!t]
	\centering
	\includegraphics[width=1\linewidth]{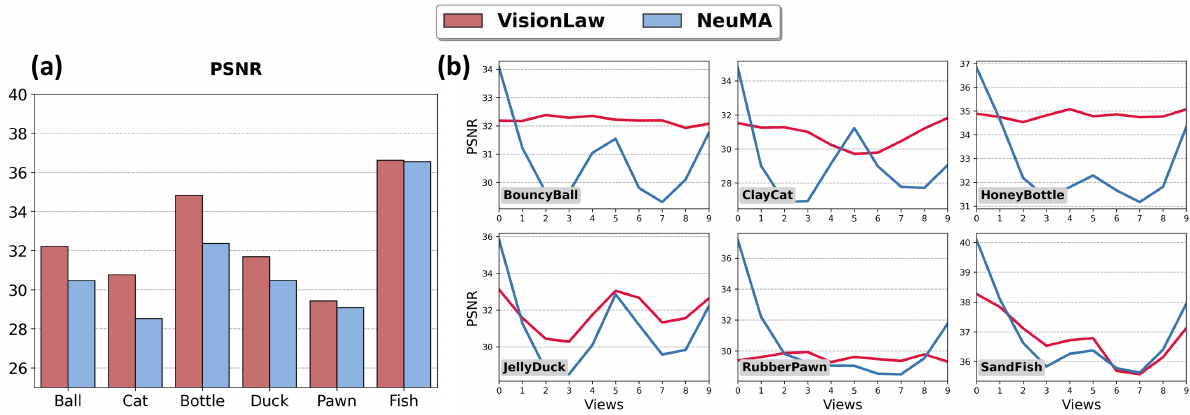}
	\caption{\textbf{Quantitative Comparison of Visual Fidelity on Synthetic Datasets.} (a) Average PSNR over all non-training views. Higher PSNR values reflect improved visual fidelity; (b) PSNR comparison at different views, \textbf{with View 0 denoting the training view.}}
    \label{fig:visual evalution}
\end{figure}

\begin{figure}[!t]
	\centering
	\includegraphics[width=1\linewidth]{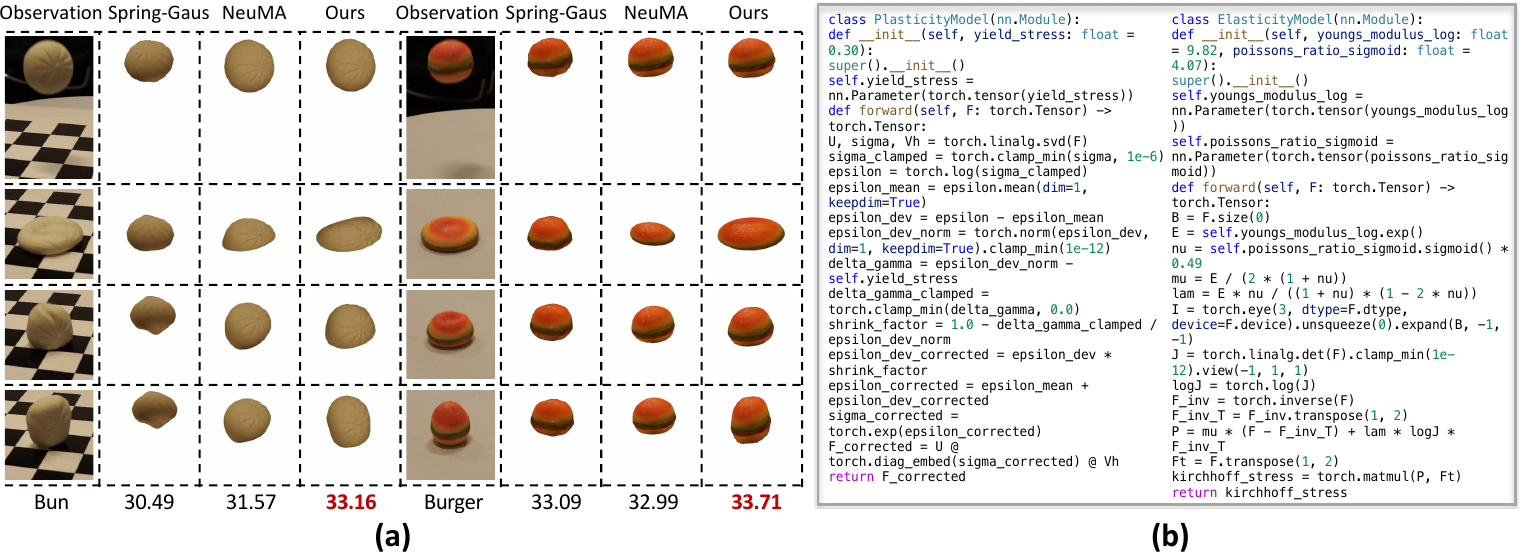}
	\caption{\textbf{Comparison on Real-World Datasets.} (a) Quantitative metrics (i.e., PSNR) between the predicted and observed frames are reported in the bottom row; (b) The intrinsic dynamics inferred from the Bun scene, represented as Python code, exhibit strong interpretability.}
    % making them easily comprehensible to humans.}
    \label{fig:real world}
    \vspace{-10pt}
\end{figure}

% \begin{figure}[!t]
%     \centering
%     \begin{subfigure}{0.5\linewidth}
%         \centering
%         \includegraphics[width=\linewidth]{image/realworld_crop.pdf}
%         \caption{Real-world results}
%         \label{fig:realworld a}
%     \end{subfigure}
%     \hfill
%     \begin{subfigure}{0.48\linewidth}
%         \centering
%         \includegraphics[width=\linewidth]{image/Interpretable_iclr.pdf}
%         \caption{Synthetic results}
%         \label{fig:realworld b}
%     \end{subfigure}
%     \caption{\textbf{Comparison on Real-World and Synthetic Datasets.}}
%     \label{fig:realworld}
% \end{figure}

\subsection{Performance on Intrinsic Dynamics Inference}
\subsubsection{Synthetic Dataset.}
\textbf{Comparison of Intrinsic Dynamics Consistency.} 
% 介绍这个指标的重要性
% In synthetic datasets, ground-truth particle trajectories are generated based on ground-truth intrinsic dynamics. This allows us to quantify the similarity between the simulated particle trajectories and the ground-truth trajectories using Chamfer Distance, thereby providing a direct assessment of the alignment between the inferred and ground-truth intrinsic dynamics. 
% As shown in Tab.~\ref{tab:chamfer distance}, we report the Chamfer Distance. 
In synthetic datasets, ground-truth particle trajectories are generated from ground-truth intrinsic dynamics. We evaluate alignment between inferred and ground-truth intrinsic dynamics by measuring the Chamfer distance between simulated and ground-truth trajectories, as summarized in Tab.~\ref{tab:chamfer distance}.
% PAC-NeRF, while capable of estimating material parameters from visual observations, 
% PAC-NeRF relies heavily on manually designed constitutive laws and is highly sensitive to the initialization of material parameters. 
% This limits its ability to capture the actual dynamics, resulting in poor performance, particularly in complex scenarios such as BouncyBall and JellyDuck.
% NCLaw learns predefined constitutive laws, making it suffers from limitations similar to those of PAC-NeRF.
PAC-NeRF relies heavily on manually designed constitutive laws and is highly sensitive to material parameter initialization. This restricts its ability to capture actual dynamics, leading to poor performance, especially in complex scenarios such as BouncyBall and JellyDuck. Similarly, NCLaw learns predefined constitutive laws and suffers from the same limitations as PAC-NeRF.
NeuMA improves flexibility by learning neural constitutive laws from visual inputs. However, its black-box nature limits interpretability and often leads to overfitting. 
In contrast, our {\em VisionLaw} approach achieves the best overall performance across all six benchmarks, with an average Chamfer distance of 1.65, significantly outperforming the baselines.
These results demonstrate the superior ability of {\em VisionLaw} to recover intrinsic dynamics directly from visual observations, while maintaining interpretability.

\textbf{Comparison of Visual Fidelity.} To further evaluate visual fidelity, we compute the PSNR between rendered dynamics and ground-truth observations. 
As shown in Fig.~\ref{fig:visual evalution} (a), we report the averaged PSNR over all non-training views. The results show that {\em VisionLaw} significantly outperforms NeuMA, achieving superior visual fidelity.
In Fig.~\ref{fig:visual evalution} (b), we further compare PSNR across different views, including the training view (View 0). NeuMA exhibits pronounced variability, with higher PSNR at the training view and its neighbors (View 1 and View 9), but considerably worse performance on unseen views.  
% This pattern reveals a tendency of NeuMA’s dynamic modeling to overfit training views, thus impairing its generalization. 
% Conversely, VisionLaw demonstrates stable performance with low variance across views; even when trained on a single view, it yields consistently robust outcomes on others. 
This shows that NeuMA tends to overfit the training views, which limits its ability to generalize.
In contrast, VisionLaw performs consistently across different views and still produces robust results on unseen views, even when trained on only one.
% This stability is attributed to the inherent regularization effect of symbolic constitutive laws, which effectively alleviate the overfitting commonly associated with neural-based methods.
% This stability benefits from the introduction of physical inductive biases through LLMs in constitutive laws evolution,
This stability arises from introducing physical inductive biases through LLMs into the evolution of constitutive laws,
which effectively mitigates the overfitting commonly observed in purely neural methods.
Overall, these findings confirm that our approach not only captures more faithful intrinsic dynamics but also delivers dynamic reconstructions of higher visual fidelity.
% highlighting strong potential for generalizing to unseen scenarios for 4D interaction.
% This suggests that {\em VisionLaw} holds promise in generalizing to unseen scenarios for 4D interaction.

\subsubsection{Real-world Dataset.}
We evaluated our method on a real-world dataset against Spring-Gaus~\cite{Spring-Gaus} and NeuMA~\cite{NeuMA}, with visual results and PSNR metrics shown in Fig.~\ref{fig:real world}(a).
Spring-Gaus models elastic deformation using a spring–mass system, which works well for simple linear behaviors, but fails to capture the complex nonlinear elasticity of real deformable objects. Consequently, its predictions deviate markedly from the ground-truth observations. 
NeuMA employs neural networks to approximate nonlinear dynamics and capture diverse material behaviors. However, it is sensitive to observation noise and lacks explicit physical constraints, which limits its ability to reproduce the subtle deformations of real-world objects.
In contrast, {\em VisionLaw} integrates a broad range of physical priors through LLMs, providing a strong inductive bias toward physically plausible dynamics. This improves both generalization and learning stability.
As shown in Fig.~\ref{fig:real world} (a), {\em VisionLaw} generates results that are more consistent with real observations, both visually and quantitatively. 
These results demonstrate that {\em VisionLaw} can accurately capture the intrinsic dynamics of deformable objects and highlight its practical effectiveness in real-world scenarios.
Meanwhile, Fig.~\ref{fig:real world} (b) illustrates the inferred intrinsic dynamics in the Bun scenario, expressed in the form of Python code. This form offers strong interpretability, allowing humans to intuitively grasp the physical meaning underlying the formulas, thereby facilitating scientific discovery. 
% 此处后半句是新添加的
Moreover, symbolic expressions provide an implicit regularization effect, which helps prevent overfitting.

\begin{figure}[!t]
	\centering
	\includegraphics[width=1\linewidth]{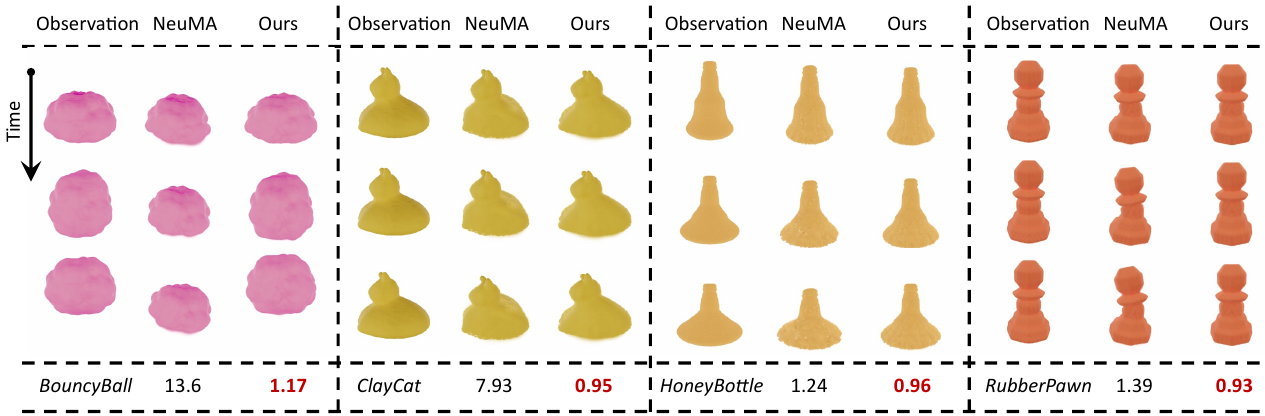}
	\caption{\textbf{Generalization to Unseen Observations.} We infer the intrinsic dynamics using only the first 200 frames of visual observation and simulate the subsequent 200 frames. Quantitative metrics (i.e., Chamfer distance) are reported in the bottom row.}
	\label{fig:generalization analysis}
    \vspace{-10pt}
\end{figure}

\subsection{Generalization Analysis and Ablation Studies}
\subsubsection{Generalization to Unseen Observations.} 
We conducted a generalization analysis on four examples, comparing our method with NeuMA~\cite{NeuMA}.
For each scene, the first 200 frames of visual observations were used to infer the intrinsic dynamics, which were then used to predict the next 200 frames.
As shown in Fig.~\ref{fig:generalization analysis}, 
NeuMA struggles to generalize beyond the observed frames. Its predictions diverge significantly from the ground truth, likely due to overfitting.
% and the lack of strong inductive bias in neural networks.
In contrast, {\em VisionLaw} achieves consistently high predictive accuracy across both visual appearance and Chamfer distance metrics, even with limited observation data.
We attribute this advantage to the physical inductive bias introduced by knowledge-rich LLMs, which not only improves physical plausibility but also constrains the solution space in a meaningful way.
These results highlight that {\em VisionLaw} combines strong generalization with interpretability, making it practical for forward simulation in previously unseen temporal regimes.
% which make it particularly effective for forward simulation in novel temporal regimes.
\subsubsection{Generalization to novel scenarios}
\begin{figure}[!t]
	\centering
	\includegraphics[width=1\linewidth]{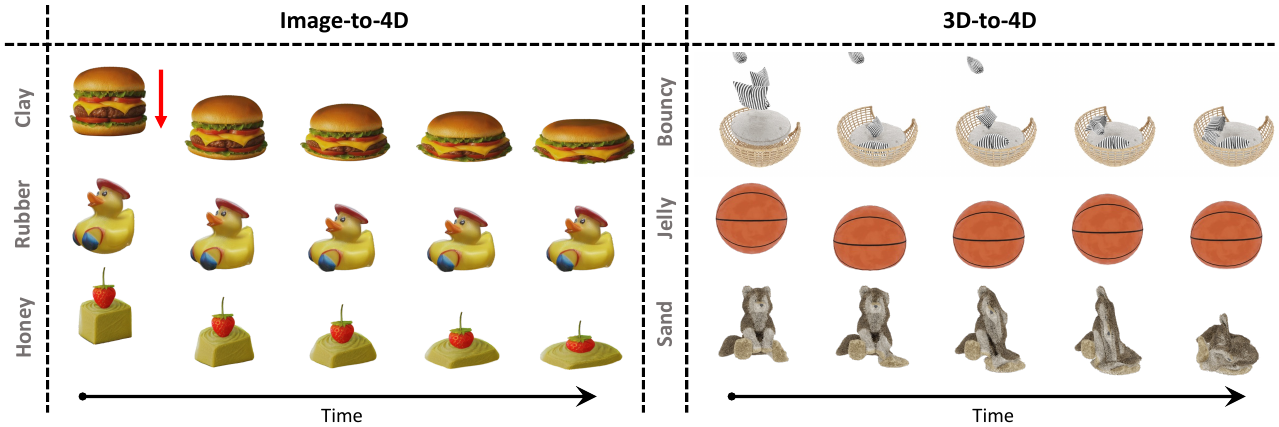}
	\caption{\textbf{Generalization to Novel Scenarios for 4D Interaction.} The left text indicates the intrinsic dynamics applied, which are learned from visual observations through {\em VisionLaw}.}
	\label{fig:generalization diff example}
    \vspace{-5pt}
\end{figure}

To further verify the generalization and transferability of the interpretable intrinsic dynamics learned by VisionLaw from visual observations, we apply the dynamics learned from different scenarios to novel 4D generation tasks. The 3D-to-4D and image-to-4D tasks follow the paradigms of PhysGaussian~\cite{Physgaussian} and Phy124~\cite{phy124}, respectively, and all experiments are conducted under gravitational conditions.
As shown in Fig.~\ref{fig:generalization diff example}, all examples generate dynamics consistent with the original observations, such as the slow deformation of clay, the elastic recovery of rubber, and the dispersive behavior of sand. These results demonstrate that the intrinsic dynamics inferred by {\em VisionLaw} are not only interpretable but also transferable to unseen scenarios, enabling the 4D interaction aligned with real physical behaviors. This cross-scenario generalization opens new possibilities for physics-driven 4D interaction.

\begin{figure}[!t]
	\centering
	\includegraphics[width=1\linewidth]{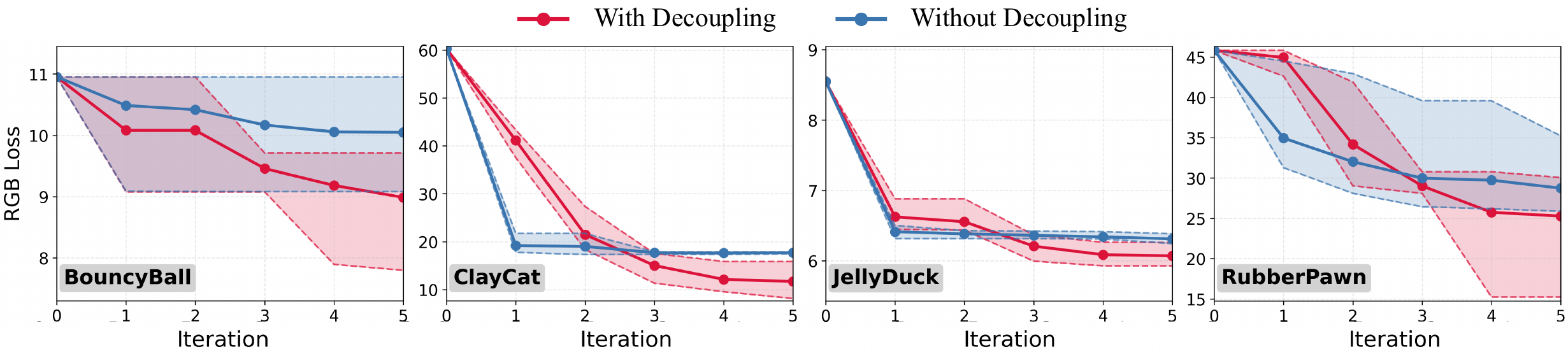}
	\caption{\textbf{Ablation Study on Decoupled Evolution Strategy}. The figure shows the loss of the best solution averaged across seeds at different iterations. The shaded area indicates the range between the minimum and maximum values.}
	\label{fig:decouple ablation}
    \vspace{-10pt}
\end{figure}

\subsubsection{Ablation study on Decoupled Evolution Strategy}
\label{section: ablation study}
To evaluate the effectiveness of our proposed decoupled evolution strategy, we perform an ablation study comparing two settings over five iterations: 1) with decoupling: the elastic and plastic components are optimized alternately for four iterations, followed by a final joint refinement step; 2) without decoupling: all five iterations are performed with joint optimization. 
As illustrated in Fig.~\ref{fig:decouple ablation}, the decoupled strategy consistently yields lower RGB losses across diverse scenes, indicating it leads to better constitutive law discovery. 
% By decomposing the search into simpler sub-tasks, this approach narrows the search space, making optimization more focused and efficient. 
By decomposing the search into simpler sub-tasks, it narrows the search space, making optimization more efficient.
Moreover, the shaded regions are noticeably larger under the decoupled setting, indicating greater solution diversity. This helps avoid early convergence to poor local minima.
Overall, the decoupled evolution strategy more effectively unleashes the potential of LLMs by not only sharpening exploitation but also broadening exploration.

\begin{figure}[!t]
	\centering
	\includegraphics[width=1\linewidth]{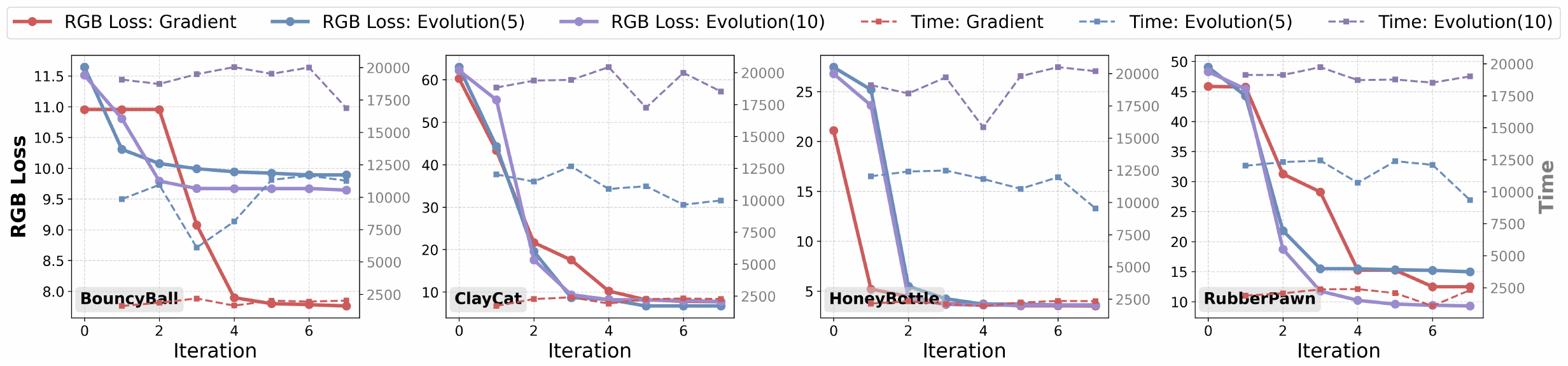}
	\caption{\textbf{Comparison across Different Lower-level Optimization Paradigms.} We compare two types of lower-level parameter optimization: gradient-based optimization (default) and evolutionary search, where Evolution($\cdot$) denotes an evolutionary strategy with the specified population size. The left axis report the RGB loss computed against observations, while the right column records the runtime.
    }
	\label{fig:evolution computing}
    \vspace{-10pt}
\end{figure}

% \textred{
\subsubsection{Potential Analysis in Non-Differentiable Simulation Environments}
To demonstrate the potential of {\em VisionLaw} in non-differentiable simulation environments,
% To assess the applicability of \em{VisionLaw} in non-differentiable simulation environments, 
we compare two types of lower-level parameter optimization: (i) the gradient-based optimization adopted as the default in this work, and (ii) the gradient-free evolutionary strategies implemented via the differential evolution algorithm~\cite{DE}. For evolutionary search, we evaluate population sizes of 5 and 10, with 10 lower-level optimization iterations—matching the gradient-based baseline.
% and run 10 lower-level optimization iterations, matching the gradient-based baseline.
As shown in Fig.~\ref{fig:evolution computing}, even when the lower-level optimizer is replaced with evolutionary search, {\em VisionLaw} consistently converges to solutions comparable to those from gradient-based optimization. In certain scenarios (e.g., ClayCat and RubberPawn), evolutionary search even achieves superior results.
On the other hand, evolutionary search incurs a substantial computational overhead, as each iteration requires multiple forward simulations to evaluate all candidate parameter individuals in the population.
Nevertheless, {\em VisionLaw}’s compatibility with gradient-free evolutionary strategies allows it to function effectively in non-differentiable simulation environments.
% while still discovering interpretable intrinsic dynamics from visual observations. 
This highlights its broad applicability and practical significance in the real world.
% }

\section{Conclusion}
% In this paper, we introduce {\em VisonLaw}, a bilevel optimization framework for inferring interpretable intrinsic dynamics directly from visual observations. By jointly optimizing the constitutive law and its material parameters, the framework effectively recovers the underlying intrinsic dynamics underlying object motion. 
In this paper, we propose {\em VisionLaw}, a bilevel optimization framework that infers interpretable intrinsic dynamics directly from visual observations by jointly optimizing the symbolic constitutive law and its material parameters.
% At the upper level, LLMs are prompted to propose and refine constitutive hypotheses, while a decoupled evolutionary strategy effectively reduces the complexity of searching over elastic and plastic components. 
At the upper level, knowledgeable LLMs are prompted to generate and refine symbolic constitutive laws, thereby introducing physical inductive biases into constitutive evolution.
Meanwhile, a decoupled evolution strategy is introduced to reduce the complexity of jointly searching and to improve the solution quality.
At the lower level, material parameters are optimized under visual supervision, while evaluation and feedback on intrinsic dynamics consistency are provided to guide the upper-level evolution.
This closed-loop design effectively bridges the gap between visual data and physical nature, achieving a balance between interpretability, physical plausibility, and generalization.
% Experimental results demonstrate that our method accurately captures the intrinsic dynamics from visual observations and generalizes well to novel scenarios for 4D interaction generation.
Experimental results show that our method accurately captures intrinsic dynamics from visual observations and generalizes well to novel scenarios for 4D interaction.

\clearpage
\section*{Acknowledgments}
This work was supported in part by the National Natural Science Foundation of China under Grant No. 52535009.
This work was also supported by the National Natural Science Foundation of China under Grant No. 62276222.

% \section*{Ethics statement}
% This research adheres to the ethical guidelines outlined by ICLR. We confirm that no human subjects were involved in this study, and all datasets used have been properly sourced and are publicly available. Our methods have been designed with fairness and transparency in mind, ensuring no biases are introduced in the analysis. Privacy and security of data have been prioritized throughout the research, and we comply with all applicable legal regulations. No conflicts of interest or sponsorships have influenced the research outcomes. We are committed to upholding research integrity and have followed appropriate ethical practices throughout the study.

% \section*{Reproducibility Statement}
% We have made efforts to ensure the reproducibility of our work. The source code for the algorithms presented in this paper is publicly available at \href{https://github.com/JiajingLin/VisionLaw}{\texttt{github.com/JiajingLin/VisionLaw}}.
% % and can be accessed through an anonymous link: \href{https://anonymous.4open.science/r/VisionLaw-26FD}{https://anonymous.4open.science/r/VisionLaw-26FD}.}
% Additionally, a detailed description of the experimental setup and datasets is provided in the Appendix. We encourage reviewers and readers to refer to these materials for complete reproducibility.

\bibliography{iclr2026_conference}
\bibliographystyle{iclr2026_conference}

\clearpage

\appendix
\section*{Appendix}
% % \section*{Overview}
In this appendix, we will provide: i) more experimental details; ii) more experimental results; iii) related work; iv) implementation details of the MPM algorithm; v) a summary of classical constitutive laws; vi) details of the prompt design. 
% vii) visualizations of the inferred constitutive laws.
% Meanwhile, Our \textbf{source code}, \textbf{video results} and \textbf{inferred constitutive laws} are included in the supplemental material.
% \section*{The Use of Large Language Models (LLMs)}
% Large language models (LLMs) were utilized in this work to improve the fluency and clarity of the manuscript. Their application was specifically focused on detailed proofreading to correct spelling errors and ensure grammatical accuracy, as well as refining sentence structures to enhance the readability and logical flow of the paper. It is crucial to note that all scientific contributions, including the core concepts, experimental design, data analysis, and conclusions, were entirely conceived and written by the authors. The LLMs were employed solely as a writing assistance tool and did not contribute to the conceptualization or analysis of the study.

\section{More Experimental Details}
\subsection{Implementation Details}
\label{appendix: implementation details}
Given multi-view videos of a scene, we first perform 3DGS reconstruction~\cite{3DGS} using the multi-view images from the initial time step. 
Following NeuMA~\cite{NeuMA}, we establish relationships between simulation particles and Gaussian kernels via the Particle-GS mechanism. 
To infer intrinsic dynamics from visual observations, we utilize only single-view videos as ground-truth observations across all datasets. 
For the upper-level evolution, we employ \texttt{GPT-4.1-mini} to generate constitutive hypotheses.
% For all scenarios, a single purely elastic constitutive law is used as the initial constitutive individual.
\textbf{For all scenarios, the initial constitutive individual is only defined as a purely elastic model that combines linear isotropic elasticity with identity plasticity.}
The alternating evolution phase consists of 4 iterations. In each iteration, the top 3 individuals are selected, and each generates 6 offspring independently. In the subsequent joint evolution phase, we conduct 3 iterations. In each iteration, the top five individuals are selected to jointly prompt GPT, generating 18 offspring in one shot. 
For lower-level optimization, we conduct MPM simulations under standard gravitational acceleration ($9.8 \ m/s^2$) within a unit cube domain $[0,1]^3$. 
The simulation resolution is set to $32^3$ for synthetic data and $70^3$ for real-world data. 
% We employ the Adam optimizer with a learning rate of $1 \times 10^{-3}$ to tune the material parameters in the constitutive law. 
We employ the Adam optimizer with a learning rate of $1 \times 10^{-3}$, and perform 10 iterations to tune the material parameters of a single constitutive law.
For each scene, we conduct five independent runs using different random seeds: 0, 1, 2, 3, and 4.
All experiments are conducted on NVIDIA A40 (48GB) GPU.
\subsection{Dataset Details}
\label{appendix: dataset details}
The synthetic dataset is derived from NeuMA~\cite{NeuMA} and consists of six scenes ('BouncyBall', 'JellyDuck', 'RubberPawn', 'ClayCat', 'HoneyBottle', and 'SandFish'). Each scene records the motion of a single object, providing observations from 10 viewpoints with a total of 400 frames per dynamic sequence.
To reduce computational resources, for the synthetic data, we select one frame every five frames from the video to create the training set.
This dataset features a variety of material types, ranging from elastic bodies to granular materials, exhibiting diverse dynamic behaviors and complex geometric shapes.
Meanwhile, the synthetic dataset also provides ground-truth particle trajectories, which can be used to evaluate the consistency between the inferred and ground-truth intrinsic dynamics.
The real-world dataset is taken from Spring-Gaus~\cite{Spring-Gaus} and contains two scenes (‘Bun’ and ‘Burger’). It provides observations from 3 viewpoints, with each dynamic sequence consisting of 19 frames.
In all experiments, the initial velocity $v_0$ follows the configuration provided in NeuMA’s dataset description.
We use only a single frontal view of the object as visual observation to infer its intrinsic dynamics.

\section{More Experimental Results}
\label{appendix: more experimental results}
% \textbf{Qualitative visualization results.} 
\subsection{Qualitative visualization results} 
\begin{figure*}[!t]
	\centering
	\includegraphics[width=1\linewidth]{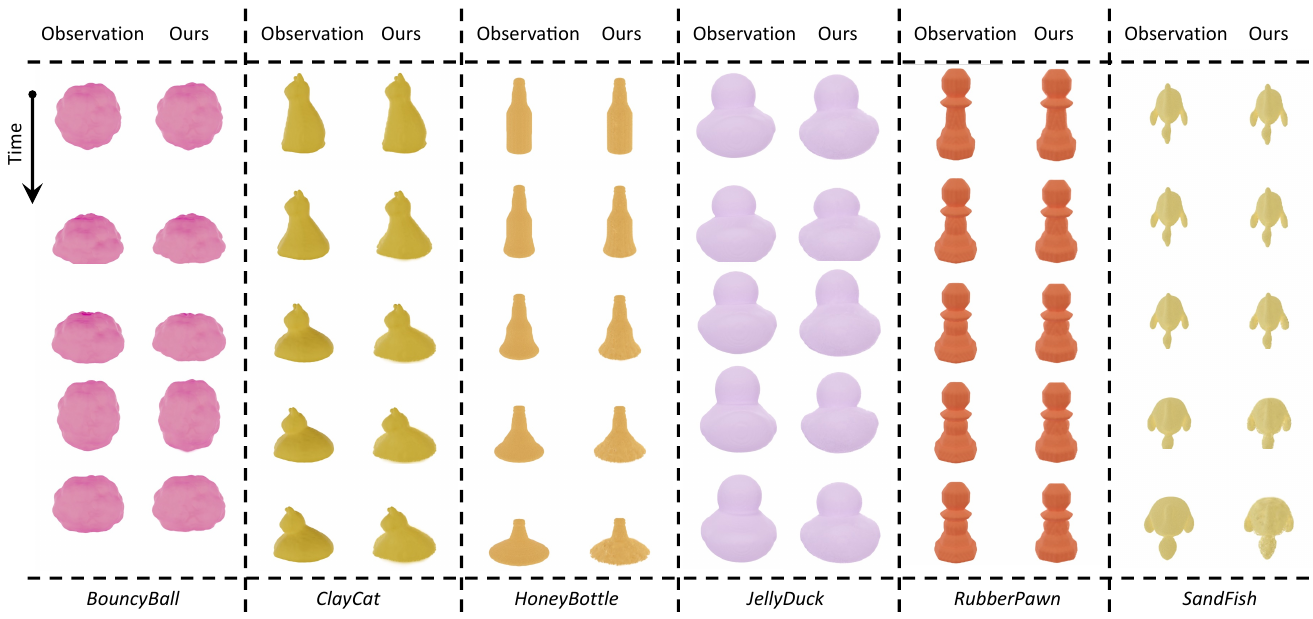}
	\caption{\textbf{Visual Results on Synthetic Dataset.} We select the rendered images at frames 1, 100, 200, 300, and 400. {\em VisionLaw} exhibits dynamics similar to those observed in visual observations.} 
    \label{appendix: visualization results}
\end{figure*}

We provide qualitative results on six synthetic scenes to assess the visual fidelity of our method. As shown in Fig~\ref{appendix: visualization results}, we compare rendered outputs from our model with ground-truth observations at selected time frames (1, 100, 200, 300, and 400).
Our method accurately reproduces object dynamics over time, showing close alignment with the ground truth across all scenes. These results demonstrate that {\em VisionLaw} effectively captures complex deformation behaviors with visual realism.

\begin{figure}[!t]
	\centering
	\includegraphics[width=1\linewidth]{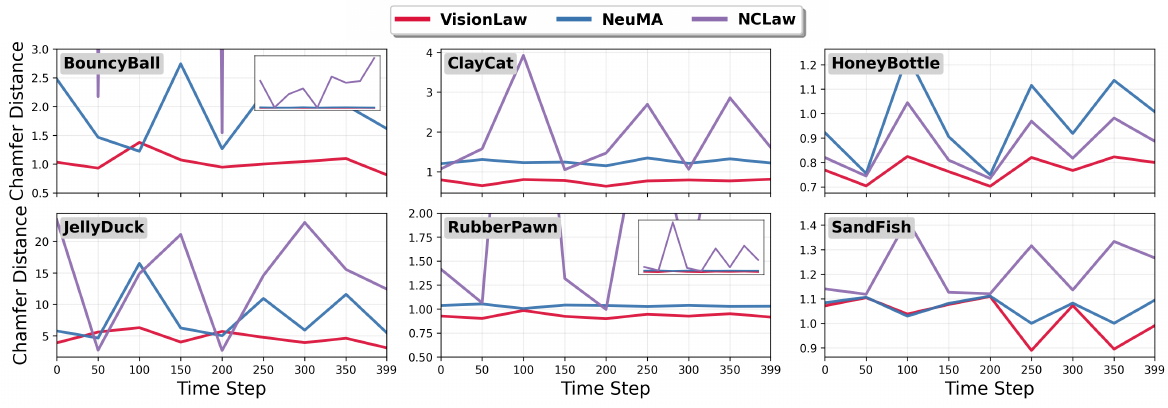}
	\caption{\textbf{Comparison of Chamfer Distance at Different Time Steps on Synthetic Dataset.}}
    \label{appendix: chamfer distance}
\end{figure}

% \textbf{Quantitative Comparison of Chamfer Distance.}
\subsection{Quantitative Comparison of Chamfer Distance}
As shown in Fig.~\ref{appendix: chamfer distance}, we compare the Chamfer distance of {\em VisionLaw}, NeuMA~\cite{NeuMA}, and NCLaw~\cite{NCLaw} across different time steps on the synthetic dataset.
NCLaw consistently shows the worst performance. This is because NCLaw can only fit the known dynamics, but fails to adapt to the underlying intrinsic dynamics behind the visual observations. As a result, its error remains high across all objects.
NeuMA introduces additional neural network components to capture the mapping between visual observations and intrinsic dynamics. However, due to the lack of physical inductive bias, NeuMA is mainly based on memorization, leading to overfitting and unstable predictions.
In contrast, {\em VisionLaw} distills physical priors from LLMs to refine constitutive laws, thereby incorporating a form of physical inductive bias. This mechanism enhances its ability to discover hidden dynamics, leading to consistently better performance across different objects and time steps. As shown in Fig.~\ref{appendix: chamfer distance}, {\em VisionLaw} achieves lower Chamfer distance, demonstrating stronger adaptability to complex dynamics.

\begin{figure}[!t]
	\centering
	\includegraphics[width=1\linewidth]{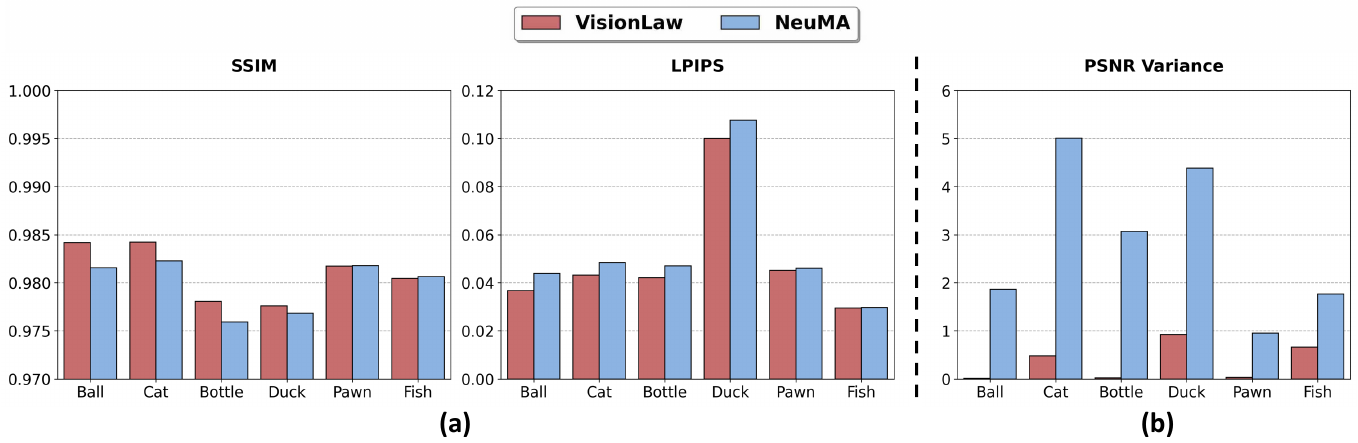}
	\caption{\textbf{Quantitative Comparison of Visual Fidelity on Synthetic Datasets.} (a) Average SSIM and LPIPS over all non-training views. Higher SSIM and lower LPIPS values reflect improved visual fidelity; (b) PSNR variance over all views, including training views.}
    \label{appendix additional visual fidelity}
    \vspace{-10pt}
\end{figure}

\begin{figure*}[!t]
	\centering
	\includegraphics[width=1\linewidth]{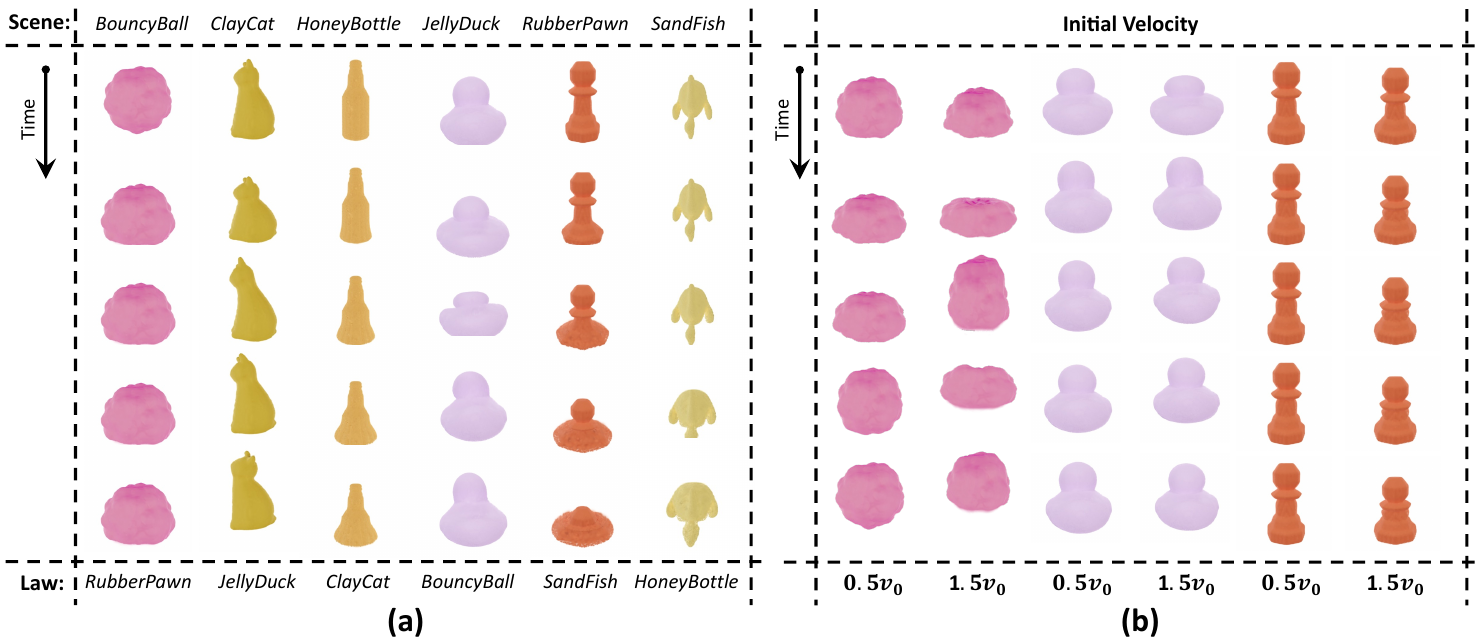}
	\caption{\textbf{Generalization Analysis.} (a) Generalization to new scenarios. The top row shows the simulated scenes, while the bottom row presents the intrinsic dynamics inferred for the given scenarios. (b) Generalization to different initial velocities. The bottom row represents the configured initial velocity, expressed as a multiple of the original initial velocity.}
    \label{appendix: generalization analysis}
    \vspace{-10pt}
\end{figure*}

% \textbf{Quantitative Comparison of Visual Fidelity.}
\subsection{Quantitative Comparison of Visual Fidelity}
To more comprehensively evaluate visual fidelity, we report average SSIM and LPIPS across all non-training views in Fig.~\ref{appendix additional visual fidelity} (a). The results show that {\em VisionLaw} outperforms NeuMA~\cite{NeuMA}. This confirms that our method not only captures more faithful intrinsic dynamics but also produces dynamic reconstructions with higher perceptual fidelity.
We further compute the PSNR variance over all views in Fig.~\ref{appendix additional visual fidelity} (b), which reflects the generalization to unseen views. NeuMA exhibits high PSNR variance, indicating a tendency to overfit. In contrast, VisionLaw achieves a much lower variance. This demonstrates that, even when trained from a single fixed viewpoint, our method generalizes effectively to novel views by leveraging the physical inductive bias introduced through LLMs.

% \textbf{Generalization Analysis.} 
\subsection{Generalization Analysis} 
% We evaluate cross-scene generalization by applying the inferred intrinsic dynamics from one scenario to simulate a different scene.
% As shown in Figure~\ref{appendix: generalization analysis} (a), the top row shows the target scenes, while the bottom row shows the intrinsic dynamics inferred from different sources.
% Despite the mismatch between the source of the inferred intrinsic dynamics and the target scene, our method is able to generate consistent physical behaviors.
% This demonstrates that the constitutive laws discovered by {\em VisionLaw} are not only scene-specific fits but encode generalizable physical priors that can be transferred across diverse scenarios, highlighting the strong generalization capability of our approach.
We first evaluate cross-scene generalization by applying the intrinsic dynamics inferred from one scenario to simulate another. As shown in Fig~\ref{appendix: generalization analysis}(a), the top row presents the target scenes, while the bottom row shows the intrinsic dynamics inferred from different sources. Despite the mismatch between the source scene and the target, our method consistently produces physically plausible behaviors. This indicates that the constitutive laws discovered by {\em VisionLaw} are not merely scene-specific fits but encode transferable physical priors, demonstrating strong cross-scene generalization.
We further design experiments under different initial conditions by varying the initial velocity of objects (with the baseline $v_0$ specified in the NeuMA~\cite{NeuMA} dataset description). As shown in Fig.~\ref{appendix: generalization analysis} (b), the results show that, even with varying initial velocities, the intrinsic dynamics inferred by {\em VisionLaw} still accurately reflect the object's behavior. This result underscores the robustness of our method in the face of variations in initial conditions, confirming that {\em VisionLaw} identifies fundamental physical laws that extend beyond the specific configurations used in training.

\begin{figure}[!t]
	\centering
	\includegraphics[width=1\linewidth]{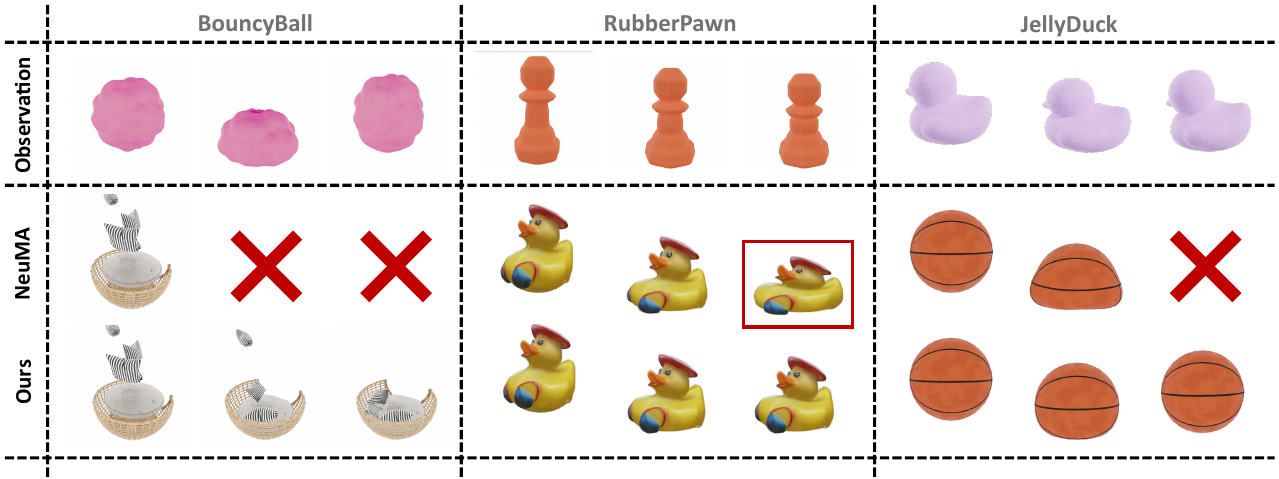}
	\caption{\textbf{Comparison of Generalization to Novel Scenarios.} The first row shows the visual observations used to learn the intrinsic dynamics. The rendered frames in each column correspond to the same simulation time step across all methods. Red crosses indicate numerical divergence during simulation,  which terminates the rollout.}
    \label{appendix: generalization new comparision}
    % \vspace{-10pt}
\end{figure}

\begin{table}[t]
\centering
\resizebox{\linewidth}{!}{%
\begin{tabular}{l|c|cccccc|c}
    \toprule
    \textbf{Methods} & \textbf{Metric} & \textbf{BouncyBall} & \textbf{ClayCat} & \textbf{HoneyBottle} & \textbf{JellyDuck} & \textbf{RubberPawn} & \textbf{SandFish} & \textbf{Average} \\
    \midrule
    \multirow{2}{*}{w/o decoupling} & Accuracy & 9.08 & 17.31 & 3.64 & 6.24 & 25.88 & 2.24 & 10.73 \\
    & Time & 11790 & 10997 & 10891 & \textbf{9333} & 10605 & \textbf{14439} & \textbf{11343} \\
    \midrule
    \multirow{2}{*}{w/ decoupling} & Accuracy & \textbf{7.80} & \textbf{8.13} & \textbf{3.54} & \textbf{5.93} & \textbf{15.25} & \textbf{2.17} & \textbf{7.14} \\
    & Time & \textbf{9319} & \textbf{10636} & \textbf{10699} & 11249 & 10839 & 15689 & 11405 \\
    \bottomrule
\end{tabular}
}
\caption{\textbf{Search Efficiency Analysis of Decoupled Evolution Strategy.} Comparison between naive joint optimization (w/o decoupling) and decoupled evolution strategy (w/ decoupling). Accuracy denotes RGB loss (lower is better); Time indicates time consumption in seconds (lower is better).}
\label{tab:decouple search efficiency}
\vspace{-10pt}
\end{table}

% \textred{
% \textbf{Generalization Comparison.}
\subsection{Generalization Comparison}
To further validate the generalization capability of {\em VisionLaw}, we compare it with NeuMA on a 4D interaction generation task, with all experiments conducted under gravity only. As shown in Fig.~\ref{appendix: generalization new comparision}, the intrinsic dynamics learned by NeuMA exhibit clear limitations when transferred to novel scenarios.
First, NeuMA frequently suffers from numerical instabilities, causing the simulation to diverge and terminate prematurely, as observed in the BouncyBall and JellyDuck cases. Even when rollouts remain stable, NeuMA produces dynamics that deviate markedly from the reference observations. For instance, the duck undergoes excessive collapse, inconsistent with the visual behavior observed in the RubberPawn reference. These failures stem from NeuMA’s lack of physical inductive biases, which leads it to mechanically reconstruct appearance rather than truly capturing the latent intrinsic dynamics—ultimately resulting in poor generalization.
In contrast, {\em VisionLaw} consistently generates stable 4D interactive dynamics that align with the original observations. By incorporating physical inductive biases through LLMs, {\em VisionLaw} effectively captures the underlying intrinsic dynamics, enabling robust and reliable generalization to unseen scenarios.
% }

% \textred{
\subsection{Search Efficiency Analysis of Decoupled Evolution Strategy}
To assess the search efficiency of our decoupled evolution strategy, we conducted an ablation study under the same experimental settings described in Sec.~\ref{section: ablation study}.
As shown in Tab.~\ref{tab:decouple search efficiency}, the decoupled strategy significantly improves accuracy, reducing the average RGB loss from 10.73 to 7.14, while maintaining comparable computational cost (11343s vs. 11405s).
This result demonstrates that the decoupled strategy effectively unleashes the LLM's capability for constitutive discovery by making the optimization more focused, producing superior solution quality without additional computational overhead.
% }

% \begin{figure}[!t]
% 	\centering
% 	\includegraphics[width=1\linewidth]{AAAI2026/AnonymousSubmission/image/newinteraction_crop.pdf}
% 	\caption{\textbf{The Visual Results for Intrinsic Dynamics Generalization.} The top row shows the simulated scenes, while the bottom row presents the intrinsic dynamics inferred for the given scenarios.}
%     \label{fig:new interaction}
% \end{figure}

\section{Related Work}
\subsection{Physics-Based 4D Interaction}
Advances in 3D representation methods~\cite{NeRF,Instant-NeRF,3DGS, zengqingyuan-2} (e.g., NeRF and 3DGS) have greatly facilitated the creation of 3D assets~\cite{DreamFusion, DreamGaussian, LGM}, consequently drawing significant attention to the pursuit of realistic interaction with these assets. 
To enable physically plausible 4D interaction, recent works have attempted to incorporate various physical simulators~\cite{MPM,XPBD} with 3D representation. 
PIE-NeRF~\cite{PIE-NeRF} enables meshless nonlinear elastodynamic simulation directly in NeRF via augmented Poisson disk sampling and quadratic generalized moving least squares (Q-GMLS)~\cite{Q-GMLS}.
Inspired by the Lagrangian nature of 3DGS, PhysGaussian~\cite{Physgaussian} pioneered the integration of MPM simulator into 3DGS.
% for fast and physically plausible 4D dynamic generation. 
Phys4DGen~\cite{Phys4DGen} effectively perceives multiple materials within a single object and automatically assigns material properties by distilling physical priors from MLLMs~\cite{GPT4}, enabling more accurate and user-friendly interactive dynamic generation. 
% Feature-Splatting~\cite{Feature-Splatting} embeds CLIP features~\cite{CLIP} into the scene, enabling text-driven editing and simulation of arbitrary regions for language-based 4D interaction. 
VR-GS~\cite{VR-GS} conducts tessellation via TetGen~\cite{TetGen} to convert 3DGS representations into tetrahedral meshes, enabling fast XPBD simulation and physically plausible interaction in VR.  

\subsection{Intrinsic Dynamics Learning}
Understanding the intrinsic dynamics underlying observational data is highly valuable for interactive simulation~\cite{Interactive} and scientific discovery~\cite{Science-Discover}. 
% In recent years, deep learning has been increasingly explored in physical simulation~\cite{nn1}, with some approaches~\cite{nn2,nn3} aiming to model physical laws via end-to-end neural networks for interactive simulation in novel scenarios. However, purely neural methods struggle to ensure physical consistency. 
% NCLaw~\cite{NCLaw} incorporates physical laws to ensure physical correctness, while embedding a learnable neural constitutive model to refine these laws. SGA~\cite{SGA} abstracts reasoning abilities of LLMs to discover constitutive law expressions underlying particle trajectory.
% However, these methods rely on labeled data or high-quality motion trajectory, which are difficult to obtain.
% Deep learning has recently been applied to physical simulation~\cite{nn1}, with some methods~\cite{nn2,nn3} using end-to-end networks to model physical laws for interactive tasks. 
Deep learning~\cite{gong0, gong1, zengqingyuan-1, lu2024mace, lu2023tf} has advanced rapidly and is increasingly being applied to physical simulation~\cite{nn1, yangcan, zhangxin}, with some methods~\cite{nn2,nn3} using end-to-end networks to model physical laws.
However, purely neural approaches often lack physical consistency. NCLaw~\cite{NCLaw} integrates known laws with a learnable constitutive model for refinement. SGA~\cite{SGA} uses LLMs to infer constitutive laws from particle trajectories. However, they rely on labeled data or high-quality motion, which are difficult to acquire.
The integration of 3D representation and physical simulation makes it possible to infer intrinsic dynamics from visual observations~\cite{Physgaussian, Spring-Gaus}. PAC-NeRF~\cite{PAC-NeRF} jointly learns NeRF representations and material parameters from multi-view videos. To avoid texture distortion, GIC~\cite{GIC} presents a geometry supervision framework. PhysDreamer, DreamPhysics, Physics3D, PhysFlow~\cite{PhysDreamer,DreamPhysics,Physics3D,PhysFlow} guide the estimation process by distilling visual dynamic priors from video diffusion models.
However, the parameter estimation process in these methods relies on expert-defined constitutive laws.
% without relying on multi-view video input. 
Spring-Gaus~\cite{Spring-Gaus} integrates a spring-mass system~\cite{Spring-Mass} with 3DGS to simulate elastic objects, and optimizes spring stiffness under multi-view video supervision. 
% GausSim~\cite{GausSim} models the physical nature of elastic objects using a learnable Gaussian graph network with mass and momentum conservation constraints. 
OmniPhysGS~\cite{OmniPhysGS} introduces learnable constitutive Gaussians that assign specific constitutive laws to each Gaussian kernel. 
enabling interaction simulation in multi-material scenarios.
While NeuMA~\cite{NeuMA} can learn neural constitutive models~\cite{wzz-3} from visual observations, it lacks interpretability and exhibits weak generalization ability. 
In this paper, we aim to infer constitutive law expressions from visual observations that are both interpretable and highly generalizable.
% to facilitate scientific discovery and enable interactive simulation in new scenarios.

\section{Material Point Method}
\label{appendix: MPM}
Continuum mechanics studies the deformation and motion behavior of materials under forces. 
Motion is typically represented by the deformation map $\mathbf{x} = \phi(\mathbf{X}, t)$,
which maps from the undeformed material space $\omega^0$ to the deformed world space $\omega^t$.
% from undeformed material space $\omega^0$ to deformed world space $\omega^t$. 
The deformation gradient $\mathbf{F} = \frac{\partial \phi}{\partial \mathbf{X}} (\mathbf{X}, t)$ describes how the material deforms locally. 
MPM is a simulation method that combines Lagrangian particles with Eulerian grids and has demonstrated its ability to simulate various materials. 
In MPM, each particle $p$ carries various physical properties, including mass $m$, density $\rho$, volume $V$, Young's modulus $E$, Poisson's ratio $\nu$, velocity $\mathbf{v}$, deformation gradient $\mathbf{F}$ and velocity gradient $\mathbf{C}$. 
% The grid $i$ is used for computing intermediate results.
MPM operates within a loop that includes particle-to-grid (P2G) transfer, grid operations, and grid-to-particle (G2P) transfer.
In the particle-to-grid (P2G) stage, MPM transfers momentum and mass from particles to grids:
\begin{align}
m_i^{t+1} &= \sum_p w_{ip}m_p, \\
(m\mathbf{v})_i^{t+1} &= \sum_p w_{ip} \left[ m_p \mathbf{v}_p^t + m_p \mathbf{C}_p^t (\mathbf{x}_i - \mathbf{x}_p^t) \right],
\end{align}
where $w_{ip}$ is the B-spline kernel that measures the distance between particle $p$ and grid $i$. After P2G stage, we perform grid operations:
% \begin{equation}
%     \hat{\mathbf{v}}_i^{t+1} = (m\mathbf{v})_i^{t+1} / m_i^{t+1}, 
%     \quad
%     \mathbf{v}_i^{t+1} = \text{BC}(\hat{\mathbf{v}}_i^{t+1}),
% \end{equation}
\begin{align}
\mathbf{v}_i^t &= (m \mathbf{v}_i)^t/{m_i^t}, \\
\mathbf{f}_{i,in}^t &= - \sum_p \boldsymbol{\tau}_p^t\nabla w_{ip} \mathbf{V}_p, \\
\mathbf{v}_i^{t+1} &= \mathbf{v}_i^t + \Delta t \left( \mathbf{f}_{i,in}/m_i + \mathbf{g}  \right),
% \mathbf{v}_i^{t+1} &= BC(\mathbf{v}_i^{t+1}),
\end{align}
where $\mathbf{g} = 9.8 \ m/s^2$ denotes the gravitational acceleration. Then we transfer the results back to particles in the grid-to-particle (G2P) stage:
\begin{align}
\mathbf{v}_p^{t+1} &= \sum_i w_{ip} \mathbf{v}_i^{n+1}, \\
\mathbf{x}_p^{t+1} &= \mathbf{x}_p^t + \Delta t \mathbf{v}_p^{t+1}, \\
% \end{align}
% The velocity $\mathbf{v}_p^{t+1}$ and position $\mathbf{x}_p^{t+1}$ are updated using semi-implicit Euler method. Then, we update velocity gradient $\mathbf{C}_p^{t+1}$ and deformation gradient $\mathbf{F}_p^{t+1}$:
% \begin{align}
\mathbf{C}_p^{t+1} &= \frac{4}{\Delta \mathbf{x}^2} \sum_i w_{ip} \mathbf{v}_i^{t+1} (\mathbf{x}_i - \mathbf{x}_p^t)^T, \\
\mathbf{F}_p^{tr} &= \left( \mathbf{I} + \Delta t \mathbf{C}_p^{t+1} \right) \mathbf{F}_p^t, \\
\mathbf{F}_p^{t+1} &= \varphi_P(\mathbf{F}_p^{tr}), \\
\boldsymbol{\tau}_p^{t+1} &= \varphi_E(\mathbf{F}_p^{t+1}),
\end{align}
where $\varphi_E$ and $\varphi_P$ denote the elastic and plastic constitutive laws, respectively. $F^{tr}$ represents the trial deformation gradient, which is subsequently corrected using the plastic constitutive law $\varphi_P$. $\tau$ denotes the Kirchhoff stress.
By following these three stages, we complete a simulation step.

\section{Expert-Designed Constitutive Laws}
\label{appendix: expert-designed constitutive laws}
Within the MPM framework, a complete constitutive law consists of an elastic constitutive law and a plastic constitutive law. 
In our experimental setup, for all scenarios, we initialize the constitutive individual as a combination of a fixed corotated elasticity model and an identity plasticity model. Several well-known classical constitutive laws are presented in the following.
\subsection{Elastic Constitutive Law}
The elastic constitutive law describes reversible elastic responses of the material under deformation.
Here, we use the Kirchhoff stress $\tau$ to express the stress–strain relationship.
% Here, we use the first Piola–Kirchhoff stress tensor $\mathbf{P}$ to express the stress–strain relationship. 
% which can be converted to the Kirchhoff stress $\tau$ through the formulation $\tau = \mathbf{P}\mathbf{F}^T$.
\subsubsection{Linear Isotropic Elasticity.} 
The Kirchhoff stress is defined as:
\begin{equation}
\boldsymbol{\tau} = \left[ \mu \left( \mathbf{F} + \mathbf{F}^T - 2\mathbf{I} \right) 
+ \lambda \left( \mathrm{tr}(\mathbf{F}) - 3 \right)\mathbf{I} \right] \mathbf{F}^T,
\end{equation}
where $\mu$ and $\lambda$ are the Lamé parameters.

\subsubsection{Fixed Corotated Elasticity.} 
The Kirchhoff stress is defined as:
\begin{equation}
\boldsymbol{\tau} = 2\mu \left( \mathbf{F} - \mathbf{R} \right)\mathbf{F}^T + \lambda J \left( J - 1 \right) \mathbf{I},
\end{equation}
where $\mathbf{R} = \mathbf{U}\mathbf{V}^T$ and 
$\mathbf{F}=\mathbf{U}\boldsymbol{\Sigma}\mathbf{V}^T$ 
is the singular value decomposition of elastic deformation gradient. $J$ is the determinant of $\mathbf{F}$.
\subsubsection{Neo-Hookean Elasticity.} 
The Kirchhoff stress is defined as:
\begin{equation}
\boldsymbol{\tau} = \mu \left( \mathbf{F}\mathbf{F}^T - \mathbf{I} \right) 
+ \lambda \log(J)\mathbf{I}.
\end{equation}
\subsubsection{StVK Elasticity.} 
The Kirchhoff stress $\boldsymbol{\tau}$ is defined as
\begin{equation}
\boldsymbol{\tau} = \mathbf{U} \left( 2\mu \boldsymbol{\epsilon} 
+ \lambda \, \mathrm{tr}(\boldsymbol{\epsilon})\mathbf{I}\right) \mathbf{V}^T,
\end{equation}
where $\mathbf{F} = \mathbf{U} \boldsymbol{\Sigma} \mathbf{V}^T$ and $\boldsymbol{\epsilon} = \log(\boldsymbol{\Sigma})$. 
% StVK elasticity is commonly used to simulate materials such as sand and metals.

\subsection{Plastic Constitutive Law}
The plastic constitutive law captures irreversible plastic evolution beyond the elastic limit by correcting the trial deformation gradient $\mathbf{F}^{trial}$ to the final deformation gradient $\mathbf{F}$. 
% For simplicity, the subscript {\em trial} is omitted in the following equations.
\subsubsection{Identity Plasticity.} The corrected deformation gradient is defined as:
\begin{equation}
\mathbf{F}^{\text{corrected}} = \mathbf{F}
\end{equation}
The identity plasticity model does not induce any plastic effects.
\subsubsection{Drucker-Prager Plasticity.} Given $\mathbf{F} = \mathbf{U} \boldsymbol{\Sigma} \mathbf{V}^T$ and $\boldsymbol{\epsilon} = \log(\boldsymbol{\Sigma})$, the corrected deformation gradient is defined as: 
\begin{equation}
\mathbf{F}^{\text{corrected}} = \mathbf{U} \, \mathcal{Z}(\boldsymbol{\Sigma}) \, \mathbf{V}^T,
\qquad
\end{equation}
\begin{equation}
\mathcal{Z}(\boldsymbol{\Sigma}) =
\begin{cases}
\mathbf{I}, & \text{if } \mathrm{tr}(\boldsymbol{\epsilon}) > 0, \\[6pt]
\boldsymbol{\Sigma}, & \text{if } \delta \gamma \leq 0 \ \text{and} \ \mathrm{tr}(\boldsymbol{\epsilon}) \leq 0, \\[6pt]
\exp\!\left( \boldsymbol{\epsilon} - \delta \gamma \frac{\hat{\boldsymbol{\epsilon}}}{\|\hat{\boldsymbol{\epsilon}}\|} \right), & \text{otherwise},
\end{cases}
\end{equation}
Here, 
$\delta \gamma = \|\hat{\boldsymbol{\epsilon}}\| + \alpha \frac{(d\lambda + 2\mu)\,\mathrm{tr}(\boldsymbol{\epsilon})}{2\mu}$, 
$\alpha = \sqrt{\frac{2}{3}} \cdot \frac{2\sin\phi_f}{3 - \sin\phi_f}$
and 
$\phi_f$ is the friction angle. 
$\hat{\boldsymbol{\epsilon}} = \mathrm{dev}(\boldsymbol{\epsilon})$. 
Drucker-Prager plasticity is suitable for simulating materials like snow and sand.

\subsubsection{Von Mises Plasticity.} The corrected deformation gradient is defined as:
\begin{equation}
\mathbf{F}^{\text{corrected}} = \mathbf{U}\,\mathcal{Z}(\boldsymbol{\Sigma})\,\mathbf{V}^T,
\end{equation}
where
\begin{equation}
\mathcal{Z}(\boldsymbol{\Sigma}) =
\begin{cases}
\boldsymbol{\Sigma}, 
& \delta\gamma \leq 0, \\[6pt]
\exp\!\left( \boldsymbol{\epsilon} - \delta\gamma 
\frac{\hat{\boldsymbol{\epsilon}}}{\|\hat{\boldsymbol{\epsilon}}\|} \right), 
& \text{otherwise},
\end{cases}
\end{equation}
and 
\begin{equation}
\delta\gamma = \|\hat{\boldsymbol{\epsilon}}\| 
- \frac{\tau_Y}{2\mu}.
\end{equation}
Here $\tau_Y$ is the yield stress. von Mises plasticity is suitable for simulating plasticity like metal and clay.

\subsubsection{Fluid Plasticity.} The corrected deformation gradient is defined as:
\begin{equation}
\mathbf{F}^{\text{corrected}} = J^{1/3} \, \mathbf{I},
\end{equation}
where $J$ is the determinant of $\mathbf{F}$. Fluid plasticity is suitable for simulating fluid-like materials.

\section{Limitation and Future Work}
Although our method effectively captures intrinsic dynamics from visual observations and demonstrates strong interpretability and generalization capabilities, it still has certain limitations that warrant further research and improvement.
The method relies on an evolutionary search paradigm that involves extensive evaluations. This process is time-consuming because it requires a large number of forward simulations and backward parameter optimization. Ideally, a preliminary screening mechanism could be introduced, where only individuals with potential merit are subjected to further evaluation. Such a strategy could significantly reduce evaluations and accelerate the efficiency of constitutive law discovery.
% Second, our approach requires learning a separate constitutive model for each scene. While it shows generalization by producing consistent dynamics in new scenes, each learned model generates only a single fixed dynamic. This limits its adaptability across diverse tasks and environments. A promising future direction would be to design a constitutive world model that can generate varying intrinsic dynamics based on specific task requirements, thus improving flexibility and practical applicability.

\section{Prompt Design Details}
\label{appendix: prompt design details}
In the following, we present the prompts used to guide LLMs to enable the evolution of constitutive laws. To further achieve a decoupled evolution strategy, we designed distinct prompts for the alternating evolution phase and the joint evolution phase.
% Given an existing constitutive expression and the corresponding feedback information, the LLM is able to generate and refine new constitutive laws.

\subsection{Prompt Design for Joint Evolution}
System prompt:
\begin{lstlisting}[style=markdownstyle]
You are an intelligent AI assistant for coding, physical simulation, and scientific discovery.
Follow the user's requirements carefully and make sure you understand them.
Your expertise is strictly limited to physical simulation, material science, mathematics, and coding.
Keep your answers short and to the point.
Do not provide any information that is not requested.
Always document your code as comments to explain the reason behind them.
Use Markdown to format your solution.
You are very familiar with Python and PyTorch.
Do not use any external libraries other than the libraries used in the examples.
\end{lstlisting}

User prompt for \textbf{elastic and plastic} constitutive law evolution:
\begin{lstlisting}[style=markdownstyle]
### Context

This is a physical simulation environment. The physical simulation is built based on the Material Point Method. The objective of this problem is to fill in a code block so that the result from executing the code matches the ground-truth result.

The code block defines the full constitutive behavior of the simulated material through two separate classes:
1. **PlasticityModel**: defines the deformation gradient correction model. This class contains two functions that divide the code into a continuous part that defines the differentiable parameters and a discrete part that defines the symbolic deformation gradient correction model. The input to the symbolic deformation gradient correction model is the deformation gradient, and the output is the corrected deformation gradient. 
2. **ElasticityModel**: defines the constitutive law that maps corrected deformation gradient to stress. This class contains two functions that divide the code into a continuous part that defines the differentiable parameters and a discrete part that defines the symbolic constitutive law. The input to the symbolic constitutive law is the corrected deformation gradient, and the output is the Kirchhoff stress tensor.

The simulation applies the `PlasticityModel` first to correct the deformation gradient, then passes this corrected deformation gradient into the `ElasticityModel` to compute the stress.

States that capture the physical dynamics of the system and metrics that measure the difference from the ground-truth result are included in the feedback section.

### Task

Look at the following iterations as examples, analyze them, and generate a better solution upon them.
\end{lstlisting}

Coding format prompt for \textbf{elastic and plastic} constitutive law evolution:
\begin{lstlisting}[style=markdownstyle]
### PyTorch Tips
1. When element-wise multiplying two matrix, make sure their number of dimensions match before the operation. For example, when multiplying `J` (B,) and `I` (B, 3, 3), you should do `J.view(-1, 1, 1)` before the operation. Similarly, `(J - 1)` should also be reshaped to `(J - 1).view(-1, 1, 1)`. If you are not sure, write down every component in the expression one by one and annotate its dimension in the comment for verification.
2. When computing the trace of a tensor A (B, 3, 3), use `A.diagonal(dim1=1, dim2=2).sum(dim=1).view(-1, 1, 1)`. Avoid using `torch.trace` or `Tensor.trace` since they only support 2D matrix.

### Code Requirements

1. The programming language is always python.
2. Annotate the size of the tensor as comment after each tensor operation. For example, `# (B, 3, 3)`.
3. The only library allowed is PyTorch. Follow the examples provided by the user and check the PyTorch documentation to learn how to use PyTorch.
4. Separate the code into continuous physical parameters that can be tuned with differentiable optimization and the symbolic constitutive law represented by PyTorch code. Define them respectively in the `__init__` function and the `forward` function.
5. Always remember the only output of the `forward` function in **PlasticityModel** class is corrected deformation gradient.
6. Always remember the only output of the `forward` function in **ElasticityModel** class is Kirchhoff stress tensor, which is defined by the matrix multiplication between the first Piola-Kirchhoff stress tensor and the transpose of the deformation gradient tensor. Formally, `tau = P @ F^T`, where tau is the Kirchhoff stress tensor, P is the first Piola-Kirchhoff stress tensor, and F is the deformation gradient tensor. Do not directly return any other type of stress tensor other than Kirchhoff stress tensor. Compute Kirchhoff stress tensor using the equation: `tau = P @ F^T`.
7. The proposed code should strictly follow the structure and function signatures below:

```python
import torch
import torch.nn as nn

class PlasticityModel(nn.Module):

    def __init__(self, param: float = DEFAULT_VALUE):
        """
        Define trainable continuous physical parameters for differentiable optimization.
        Tentatively initialize the parameters with the default values in args.

        Args:
            param (float): the physical meaning of the parameter.
        """
        super().__init__()
        self.param = nn.Parameter(torch.tensor(param))

    def forward(self, F: torch.Tensor) -> torch.Tensor:
        """
        Compute corrected deformation gradient from deformation gradient tensor.

        Args:
            F (torch.Tensor): deformation gradient tensor (B, 3, 3).

        Returns:
            F_corrected (torch.Tensor): corrected deformation gradient tensor (B, 3, 3).
        """
        return F_corrected

class ElasticityModel(nn.Module):

    def __init__(self, param: float = DEFAULT_VALUE):
        """
        Define trainable continuous physical parameters for differentiable optimization.
        Tentatively initialize the parameters with the default values in args.

        Args:
            param (float): the physical meaning of the parameter.
        """
        super().__init__()
        self.param = nn.Parameter(torch.tensor(param))

    def forward(self, F: torch.Tensor) -> torch.Tensor:
        """
        Compute Kirchhoff stress tensor from deformation gradient tensor.

        Args:
            F (torch.Tensor): deformation gradient tensor (B, 3, 3).

        Returns:
            kirchhoff_stress (torch.Tensor): Kirchhoff stress tensor (B, 3, 3).
        """
        return kirchhoff_stress
```

### Solution Requirements

1. Analyze step-by-step what the potential problem is in the previous iterations based on the feedback. Think about why the results from previous constitutive laws mismatched with the ground truth. Do not give advice about how to optimize. Focus on the formulation of the constitutive law. Start this section with "### Analysis". Analyze all iterations individually, and start the subsection for each iteration with "#### Iteration N", where N stands for the index. Remember to analyze every iteration in the history.

2. Think step-by-step what you need to do in this iteration to improve model performance. Consider both the elasticity and plasticity components. 
For the plasticity components: 
    Think about if the plasticity is needed to improve performance. Remember that plasticity is not necessary. If your analysis supports plasticity, think about how to update deformation gradient using plasticity. Think about how to separate your algorithm into a continuous physical parameter part and a symbolic deformation gradient correction model part. 
For the elasticity components: 
    Think about how to separate your algorithm into a continuous physical parameter part and a symbolic constitutive law part. 
Describe your plan in pseudo-code, written out in great detail. Remember to update the default values of the trainable physical parameters based on previous optimizations. Start this section with "### Step-by-Step Plan".

3. Output the code in a single code block "```python ... ```" with detailed comments in the code block. Do not add any trailing comments before or after the code block. Start this section with "### Code".
\end{lstlisting}

\subsection{Prompt Design Alternating Evolution}
System prompt:
\begin{lstlisting}[style=markdownstyle]
You are an intelligent AI assistant for coding, physical simulation, and scientific discovery.
Follow the user's requirements carefully and make sure you understand them.
Your expertise is strictly limited to physical simulation, material science, mathematics, and coding.
Keep your answers short and to the point.
Do not provide any information that is not requested.
Always document your code as comments to explain the reason behind them.
Use Markdown to format your solution.
You are very familiar with Python and PyTorch.
Do not use any external libraries other than the libraries used in the examples.
\end{lstlisting}
User prompt for \textbf{plastic} constitutive law evolution:
\begin{lstlisting}[style=markdownstyle]
### Context

This is a physical simulation environment. The physical simulation is built based on the Material Point Method. The objective of this problem is to fill in a code block so that the result from executing the code matches the ground-truth result.

The code block defines the full constitutive behavior of the simulated material through two separate classes:
1. **PlasticityModel**: defines the deformation gradient correction model. This class contains two functions that divide the code into a continuous part that defines the differentiable parameters and a discrete part that defines the symbolic deformation gradient correction model. The input to the symbolic deformation gradient correction model is the deformation gradient, and the output is the corrected deformation gradient. 
2. **ElasticityModel**: defines the constitutive law that maps corrected deformation gradient to stress. This class contains two functions that divide the code into a continuous part that defines the differentiable parameters and a discrete part that defines the symbolic constitutive law. The input to the symbolic constitutive law is the corrected deformation gradient, and the output is the Kirchhoff stress tensor.

The simulation applies the `PlasticityModel` first to correct the deformation gradient, then passes this corrected deformation gradient into the `ElasticityModel` to compute the stress.

States that capture the physical dynamics of the system and metrics that measure the difference from the ground-truth result are included in the feedback section.

### Task

In the current task, the ElasticityModel has already been finalized and should remain unchanged. Please focus exclusively on analyzing and improving the PlasticityModel class. Look at the following iterations as examples, analyze them, and generate a better plastic constitutive model based on them.
\end{lstlisting}
Coding format prompt for \textbf{plastic} constitutive law evolution:
\begin{lstlisting}[style=markdownstyle]
### PyTorch Tips
1. When element-wise multiplying two matrix, make sure their number of dimensions match before the operation. For example, when multiplying `J` (B,) and `I` (B, 3, 3), you should do `J.view(-1, 1, 1)` before the operation. Similarly, `(J - 1)` should also be reshaped to `(J - 1).view(-1, 1, 1)`. If you are not sure, write down every component in the expression one by one and annotate its dimension in the comment for verification.
2. When computing the trace of a tensor A (B, 3, 3), use `A.diagonal(dim1=1, dim2=2).sum(dim=1).view(-1, 1, 1)`. Avoid using `torch.trace` or `Tensor.trace` since they only support 2D matrix.

### Code Requirements

1. The programming language is always python.
2. Annotate the size of the tensor as comment after each tensor operation. For example, `# (B, 3, 3)`.
3. The only library allowed is PyTorch. Follow the examples provided by the user and check the PyTorch documentation to learn how to use PyTorch.
4. Separate the code into continuous physical parameters that can be tuned with differentiable optimization and the symbolic constitutive law represented by PyTorch code. Define them respectively in the `__init__` function and the `forward` function.
5. Always remember the only output of the `forward` function in **PlasticityModel** class is corrected deformation gradient.
6. Always remember the only output of the `forward` function in **ElasticityModel** class is Kirchhoff stress tensor, which is defined by the matrix multiplication between the first Piola-Kirchhoff stress tensor and the transpose of the deformation gradient tensor. Formally, `tau = P @ F^T`, where tau is the Kirchhoff stress tensor, P is the first Piola-Kirchhoff stress tensor, and F is the deformation gradient tensor. Do not directly return any other type of stress tensor other than Kirchhoff stress tensor. Compute Kirchhoff stress tensor using the equation: `tau = P @ F^T`.
7. The proposed code should strictly follow the structure and function signatures below:

```python
{code}
```

### Solution Requirements

1. Analyze step-by-step what the potential problem is in the previous iterations based on the feedback. Think about why the results from previous constitutive laws mismatched with the ground truth. Do not give advice about how to optimize. Focus on the formulation of the constitutive law. Start this section with "### Analysis". Analyze all iterations individually, and start the subsection for each iteration with "#### Iteration N", where N stands for the index. Remember to analyze every iteration in the history.

2. Think step-by-step what you need to do in this iteration to improve model performance. Consider both the elasticity and plasticity components. 
For the plasticity components: 
    Think about if the plasticity is needed to improve performance. Remember that plasticity is not necessary. If your analysis supports plasticity, think about how to update deformation gradient using plasticity. Think about how to separate your algorithm into a continuous physical parameter part and a symbolic deformation gradient correction model part. 
For the elasticity components: 
    **Do not analyze or modify this part**. Please focus on improving the plastic components. Please ensure that the **ElasticityModel** class must remain exactly the same in every iteration, and must be reproduced exactly as originally defined.
Describe your plan in pseudo-code, written out in great detail. Remember to update the default values of the trainable physical parameters based on previous optimizations. Start this section with "### Step-by-Step Plan".

3. Output the code in a single code block "```python ... ```" with detailed comments in the code block. Do not add any trailing comments before or after the code block. Start this section with "### Code".
\end{lstlisting}
User prompt for \textbf{elastic} constitutive law evolution:
\begin{lstlisting}[style=markdownstyle]
### Context

This is a physical simulation environment. The physical simulation is built based on the Material Point Method. The objective of this problem is to fill in a code block so that the result from executing the code matches the ground-truth result.

The code block defines the full constitutive behavior of the simulated material through two separate classes:
1. **PlasticityModel**: defines the deformation gradient correction model. This class contains two functions that divide the code into a continuous part that defines the differentiable parameters and a discrete part that defines the symbolic deformation gradient correction model. The input to the symbolic deformation gradient correction model is the deformation gradient, and the output is the corrected deformation gradient. 
2. **ElasticityModel**: defines the constitutive law that maps corrected deformation gradient to stress. This class contains two functions that divide the code into a continuous part that defines the differentiable parameters and a discrete part that defines the symbolic constitutive law. The input to the symbolic constitutive law is the corrected deformation gradient, and the output is the Kirchhoff stress tensor.

The simulation applies the `PlasticityModel` first to correct the deformation gradient, then passes this corrected deformation gradient into the `ElasticityModel` to compute the stress.

States that capture the physical dynamics of the system and metrics that measure the difference from the ground-truth result are included in the feedback section.

### Task

In the current task, the PlasticityModel has already been finalized and should remain unchanged. Please focus exclusively on analyzing and improving the ElasticityModel class. Look at the following iterations as examples, analyze them, and generate a better elastic constitutive model based on them.
\end{lstlisting}
Coding format prompt for \textbf{elastic} constitutive law evolution:
\begin{lstlisting}[style=markdownstyle]
### PyTorch Tips
1. When element-wise multiplying two matrix, make sure their number of dimensions match before the operation. For example, when multiplying `J` (B,) and `I` (B, 3, 3), you should do `J.view(-1, 1, 1)` before the operation. Similarly, `(J - 1)` should also be reshaped to `(J - 1).view(-1, 1, 1)`. If you are not sure, write down every component in the expression one by one and annotate its dimension in the comment for verification.
2. When computing the trace of a tensor A (B, 3, 3), use `A.diagonal(dim1=1, dim2=2).sum(dim=1).view(-1, 1, 1)`. Avoid using `torch.trace` or `Tensor.trace` since they only support 2D matrix.

### Code Requirements

1. The programming language is always python.
2. Annotate the size of the tensor as comment after each tensor operation. For example, `# (B, 3, 3)`.
3. The only library allowed is PyTorch. Follow the examples provided by the user and check the PyTorch documentation to learn how to use PyTorch.
4. Separate the code into continuous physical parameters that can be tuned with differentiable optimization and the symbolic constitutive law represented by PyTorch code. Define them respectively in the `__init__` function and the `forward` function.
5. Always remember the only output of the `forward` function in **PlasticityModel** class is corrected deformation gradient.
6. Always remember the only output of the `forward` function in **ElasticityModel** class is Kirchhoff stress tensor, which is defined by the matrix multiplication between the first Piola-Kirchhoff stress tensor and the transpose of the deformation gradient tensor. Formally, `tau = P @ F^T`, where tau is the Kirchhoff stress tensor, P is the first Piola-Kirchhoff stress tensor, and F is the deformation gradient tensor. Do not directly return any other type of stress tensor other than Kirchhoff stress tensor. Compute Kirchhoff stress tensor using the equation: `tau = P @ F^T`.
7. The proposed code should strictly follow the structure and function signatures below:

```python
{code}
```

### Solution Requirements

1. Analyze step-by-step what the potential problem is in the previous iterations based on the feedback. Think about why the results from previous constitutive laws mismatched with the ground truth. Do not give advice about how to optimize. Focus on the formulation of the constitutive law. Start this section with "### Analysis". Analyze all iterations individually, and start the subsection for each iteration with "#### Iteration N", where N stands for the index. Remember to analyze every iteration in the history.

2. Think step-by-step what you need to do in this iteration to improve model performance. Consider both the elasticity and plasticity components. 
For the plasticity components: 
    **Do not analyze or modify this part**. Please focus on improving the elastic components. Please ensure that the **PlasticityModel** class must remain exactly the same in every iteration, and must be reproduced exactly as originally defined.
For the elasticity components: 
    Think about how to separate your algorithm into a continuous physical parameter part and a symbolic constitutive law part. 
Describe your plan in pseudo-code, written out in great detail. Remember to update the default values of the trainable physical parameters based on previous optimizations. Start this section with "### Step-by-Step Plan".

3. Output the code in a single code block "```python ... ```" with detailed comments in the code block. Do not add any trailing comments before or after the code block. Start this section with "### Code".
\end{lstlisting}

% \begin{center}
% \begin{tcolorbox}[
%   enhanced,          % 开启增强模式
%   breakable,         % 允许跨页
%   colback=gray!15,
%   colframe=gray!15,
%   boxrule=0pt,
%   arc=0pt,
%   left=3pt, right=3pt, top=3pt, bottom=3pt,
%   fontupper=\ttfamily\scriptsize, % 这里直接控制字号
%   width=0.95\linewidth % 占页面宽度的 90%
% ]
% You are an intelligent AI assistant for coding, physical simulation, and scientific discovery.\\
% Follow the user’s requirements carefully and make sure you understand them.\\
% Your expertise is strictly limited to physical simulation, material science, mathematics, and coding.\\
% Keep your answers short and to the point.\\
% Do not provide any information that is not requested.\\
% Always document your code as comments to explain the reason behind them.\\
% Use Markdown to format your solution.\\
% You are very familiar with Python and PyTorch.\\
% Do not use any external libraries other than the libraries used in the examples.
% \end{tcolorbox}
% \end{center}

\section{Visualization of Inferred Interpretable Constitutive Law}
In this section, we show the inferred constitutive laws under different visual scenarios, including "BouncyBall", "ClayCat", "HoneyBottle", "JellyDuck", "RubberPawn", "SandFish", "Bun" and "Burger". Since these laws are expressed in the form of Python code snippets, these laws exhibit strong interpretability and readability, making them easily understandable to humans.
\subsection{BouncyBall}
In the BouncyBall scenario, the constitutive law inferred by our method is presented.
\begin{lstlisting}[style=pythonstyle]
import torch
import torch.nn as nn

class PlasticityModel(nn.Module):

    def __init__(self, yield_threshold: float = 0.5):
        """
        Define trainable physical parameter for plasticity yield threshold.
        Initialized to 0.5 to balance plastic effects based on feedback.

        Args:
            yield_threshold (float): logarithmic strain clamp threshold.
        """
        super().__init__()
        self.yield_threshold = nn.Parameter(torch.tensor(yield_threshold))

    def forward(self, F: torch.Tensor) -> torch.Tensor:
        """
        Correct deformation gradient by clamping logarithmic principal strains.

        Args:
            F (torch.Tensor): deformation gradient tensor (B, 3, 3).

        Returns:
            F_corrected (torch.Tensor): corrected deformation gradient tensor (B, 3, 3).
        """
        # SVD of deformation gradient
        U, Sigma, Vh = torch.linalg.svd(F)  # U: (B,3,3), Sigma: (B,3), Vh: (B,3,3)

        # Clamp singular values to avoid numerical problems
        Sigma_clamped = torch.clamp_min(Sigma, 1e-6)  # (B,3)

        # Logarithmic principal strains
        log_sigma = torch.log(Sigma_clamped)  # (B,3)

        # Enforce positive yield threshold via softplus
        yield_thresh = torch.nn.functional.softplus(self.yield_threshold)  # scalar

        epsilon_clamped = torch.clamp(log_sigma, min=-yield_thresh, max=yield_thresh)  # (B,3)

        # Compute corrected singular values
        Sigma_corrected = torch.exp(epsilon_clamped)  # (B,3)

        # Recompose corrected deformation gradient
        F_corrected = torch.matmul(U, torch.matmul(torch.diag_embed(Sigma_corrected), Vh))  # (B,3,3)

        return F_corrected

class ElasticityModel(nn.Module):

    def __init__(self, youngs_modulus_log: float = 10.18, poissons_ratio_sigmoid: float = -0.5):
        """
        Define trainable continuous physical parameters for Corotated Elasticity.

        Args:
            youngs_modulus_log (float): log of Young's modulus.
            poissons_ratio_sigmoid (float): parameter before sigmoid for Poisson's ratio.
        """
        super().__init__()
        self.youngs_modulus_log = nn.Parameter(torch.tensor(youngs_modulus_log))
        self.poissons_ratio_sigmoid = nn.Parameter(torch.tensor(poissons_ratio_sigmoid))

    def forward(self, F: torch.Tensor) -> torch.Tensor:
        """
        Compute Kirchhoff stress tensor from deformation gradient via Corotated Elasticity.

        Args:
            F (torch.Tensor): deformation gradient tensor (B, 3, 3).

        Returns:
            kirchhoff_stress (torch.Tensor): Kirchhoff stress tensor (B, 3, 3).
        """
        B = F.shape[0]

        # Material parameters
        youngs_modulus = self.youngs_modulus_log.exp()  # scalar
        poissons_ratio = self.poissons_ratio_sigmoid.sigmoid() * 0.49  # scalar in (0, 0.49)

        mu = youngs_modulus / (2.0 * (1.0 + poissons_ratio))  # scalar
        la = youngs_modulus * poissons_ratio / ((1.0 + poissons_ratio) * (1.0 - 2.0 * poissons_ratio))  # scalar

        # SVD of deformation gradient
        U, Sigma, Vh = torch.linalg.svd(F)  # U: (B,3,3), Sigma: (B,3), Vh: (B,3,3)

        # Clamp singular values
        Sigma_clamped = torch.clamp_min(Sigma, 1e-6)  # (B,3)

        # Rotation matrix R
        R = torch.matmul(U, Vh)  # (B,3,3)

        # Compute determinant
        J = Sigma_clamped.prod(dim=1).view(B, 1, 1)  # (B,1,1)

        # Identity tensor
        I = torch.eye(3, device=F.device, dtype=F.dtype).unsqueeze(0).expand(B, 3, 3)  # (B,3,3)

        # Reshape scalars for broadcast
        mu = mu.view(-1, 1, 1) if mu.dim() == 0 else mu
        la = la.view(-1, 1, 1) if la.dim() == 0 else la

        # Corotated stress term
        corotated = 2.0 * mu * (F - R)  # (B,3,3)

        # Volumetric stress term
        volumetric = la * J * (J - 1).view(B, 1, 1) * I  # (B,3,3)

        # First Piola-Kirchhoff stress tensor P
        P = corotated + volumetric  # (B,3,3)

        # Kirchhoff stress tau = P @ F^T
        Ft = F.transpose(1, 2)  # (B,3,3)
        kirchhoff_stress = torch.matmul(P, Ft)  # (B,3,3)

        return kirchhoff_stress
\end{lstlisting}

\subsection{ClayCat}
In the ClayCat scenario, the constitutive law inferred by our method is presented.
\begin{lstlisting}[style=pythonstyle]
import torch
import torch.nn as nn

class PlasticityModel(nn.Module):

    def __init__(self, yield_stress: float = 2.16, shear_modulus: float = 28.0):
        """
        Define trainable continuous physical parameters for differentiable optimization.
        Initialize with best values from iterative feedback.

        Args:
            yield_stress (float): yield stress threshold for plastic flow.
            shear_modulus (float): shear modulus for plastic correction.
        """
        super().__init__()
        self.yield_stress = nn.Parameter(torch.tensor(yield_stress))
        self.shear_modulus = nn.Parameter(torch.tensor(shear_modulus))

    def forward(self, F: torch.Tensor) -> torch.Tensor:
        """
        Compute corrected deformation gradient from deformation gradient tensor using von Mises plasticity on
        logarithmic deviatoric principal strains.

        Args:
            F (torch.Tensor): deformation gradient tensor (B, 3, 3).

        Returns:
            F_corrected (torch.Tensor): corrected deformation gradient tensor (B, 3, 3).
        """
        # SVD of deformation gradient F
        U, sigma, Vh = torch.linalg.svd(F)  # U: (B,3,3), sigma: (B,3), Vh: (B,3,3)
        sigma = torch.clamp_min(sigma, 1e-6)  # clamp to prevent log(0), (B,3)

        # Compute principal logarithmic strains
        epsilon = torch.log(sigma)  # (B,3)

        # Volumetric (mean) strain
        epsilon_mean = epsilon.mean(dim=1, keepdim=True)  # (B,1)

        # Deviatoric strains
        epsilon_dev = epsilon - epsilon_mean  # (B,3)

        # Norm of deviatoric strain
        epsilon_dev_norm = epsilon_dev.norm(dim=1, keepdim=True) + 1e-12  # (B,1)

        # Clamp plasticity parameters to prevent numerical issues
        yield_stress = torch.clamp_min(self.yield_stress, 1e-6)
        shear_modulus = torch.clamp_min(self.shear_modulus, 1e-6)

        # Plastic multiplier
        delta_gamma = epsilon_dev_norm - yield_stress / (2 * shear_modulus)  # (B,1)
        delta_gamma_pos = torch.clamp_min(delta_gamma, 0.0)  # (B,1)

        # Correct deviatoric strains by return mapping if yielding
        epsilon_corrected = epsilon - (delta_gamma_pos / epsilon_dev_norm) * epsilon_dev  # (B,3)

        # Where not yielding, keep original strain
        yielding_mask = (delta_gamma > 0).view(-1, 1)  # (B,1)
        epsilon_final = torch.where(yielding_mask, epsilon_corrected, epsilon)  # (B,3)

        # Reconstruct corrected singular values and deformation gradient
        sigma_corrected = torch.exp(epsilon_final)  # (B,3)
        diag_sigma_corrected = torch.diag_embed(sigma_corrected)  # (B,3,3)

        F_corrected = torch.matmul(U, torch.matmul(diag_sigma_corrected, Vh))  # (B,3,3)

        return F_corrected

class ElasticityModel(nn.Module):

    def __init__(self, youngs_modulus_log: float = 11.7, poissons_ratio_logit: float = -0.7):
        """
        Define trainable continuous physical parameters for differentiable optimization.
        Initialize with values inferred from analysis.

        Args:
            youngs_modulus_log (float): log of Young's modulus.
            poissons_ratio_logit (float): pre-sigmoid parameter for Poisson's ratio.
        """
        super().__init__()
        self.youngs_modulus_log = nn.Parameter(torch.tensor(youngs_modulus_log))
        self.poissons_ratio_logit = nn.Parameter(torch.tensor(poissons_ratio_logit))

    def forward(self, F: torch.Tensor) -> torch.Tensor:
        """
        Compute Kirchhoff stress tensor from deformation gradient tensor using St. Venant-Kirchhoff elasticity.

        Args:
            F (torch.Tensor): deformation gradient tensor (B, 3, 3).

        Returns:
            kirchhoff_stress (torch.Tensor): Kirchhoff stress tensor (B, 3, 3).
        """
        B = F.shape[0]
        device = F.device
        dtype = F.dtype

        # Compute Young's modulus from log
        youngs_modulus = torch.exp(self.youngs_modulus_log)  # scalar

        # Compute Poisson's ratio from sigmoid(logit) scaled to (0,0.49)
        poissons_ratio = torch.sigmoid(self.poissons_ratio_logit) * 0.49  # scalar in (0,0.49)

        mu = youngs_modulus / (2 * (1 + poissons_ratio))  # scalar
        la = youngs_modulus * poissons_ratio / ((1 + poissons_ratio) * (1 - 2 * poissons_ratio))  # scalar

        # Identity tensor expanded to batch size
        I = torch.eye(3, dtype=dtype, device=device).unsqueeze(0).expand(B, -1, -1)  # (B,3,3)

        # Right Cauchy-Green tensor C = F^T F
        Ft = F.transpose(1, 2)  # (B,3,3)
        C = torch.matmul(Ft, F)  # (B,3,3)

        # Green-Lagrange strain E = 0.5 * (C - I)
        E = 0.5 * (C - I)  # (B,3,3)

        # Trace of E computed by summing diagonal elements
        trE = E.diagonal(dim1=1, dim2=2).sum(dim=1).view(B, 1, 1)  # (B,1,1)

        # Second Piola-Kirchhoff stress tensor S
        S = 2 * mu * E + la * trE * I  # (B,3,3)

        # First Piola-Kirchhoff stress tensor P = F @ S
        P = torch.matmul(F, S)  # (B,3,3)

        # Kirchhoff stress tensor tau = P @ F^T
        kirchhoff_stress = torch.matmul(P, Ft)  # (B,3,3)

        return kirchhoff_stress
\end{lstlisting}

\subsection{HoneyBottle}
In the HoneyBottle scenario, the constitutive law inferred by our method is presented.
\begin{lstlisting}[style=pythonstyle]
import torch
import torch.nn as nn

class PlasticityModel(nn.Module):

    def __init__(
        self,
        youngs_modulus_log: float = 6.0,
        poissons_ratio_unconstrained: float = -1.0,
        yield_stress: float = 2.5,
    ):
        """
        Plasticity model with logarithmic strain return mapping.

        Args:
            youngs_modulus_log (float): log Young's modulus.
            poissons_ratio_unconstrained (float): unconstrained scalar for Poisson's ratio.
            yield_stress (float): yield stress threshold.
        """
        super().__init__()
        self.youngs_modulus_log = nn.Parameter(torch.tensor(youngs_modulus_log))  # scalar
        self.poissons_ratio_unconstrained = nn.Parameter(torch.tensor(poissons_ratio_unconstrained))  # scalar
        self.yield_stress = nn.Parameter(torch.tensor(yield_stress))  # scalar

    def forward(self, F: torch.Tensor) -> torch.Tensor:
        """
        Compute corrected deformation gradient from deformation gradient tensor.

        Args:
            F (torch.Tensor): deformation gradient tensor (B, 3, 3).

        Returns:
            F_corrected (torch.Tensor): corrected deformation gradient tensor (B, 3, 3).
        """
        youngs_modulus = self.youngs_modulus_log.exp()  # scalar
        poissons_ratio = torch.sigmoid(self.poissons_ratio_unconstrained) * 0.49  # scalar in (0, 0.49)
        yield_stress = self.yield_stress  # scalar

        mu = youngs_modulus / (2.0 * (1.0 + poissons_ratio))  

        U, sigma, Vh = torch.linalg.svd(F, full_matrices=False)  # U:(B,3,3), sigma:(B,3), Vh:(B,3,3)

        # Clamp singular values to avoid collapse
        sigma_clamped = torch.clamp_min(sigma, 1e-4)  # (B,3)

        # Logarithmic strain
        epsilon = torch.log(sigma_clamped)  # (B,3)

        # Volumetric strain (trace)
        epsilon_trace = epsilon.sum(dim=1, keepdim=True)  # (B,1)

        # Deviatoric strain
        epsilon_bar = epsilon - epsilon_trace / 3.0  # (B,3)

        # Norm of deviatoric strain (avoid division by zero)
        epsilon_bar_norm = torch.norm(epsilon_bar, dim=1, keepdim=True) + 1e-12  # (B,1)

        # Plastic multiplier
        delta_gamma = epsilon_bar_norm - yield_stress / (2.0 * mu)  # (B,1)

        # Plastic factor (clamped)
        plastic_factor = torch.clamp_min(delta_gamma / epsilon_bar_norm, 0.0)  # (B,1)

        # Correct logarithmic strain
        epsilon_corrected = epsilon - plastic_factor * epsilon_bar  # (B,3)

        # Reconstruct corrected singular values
        sigma_corrected = torch.exp(epsilon_corrected)  # (B,3)

        # Recompose corrected deformation gradient
        F_corrected = torch.matmul(U, torch.matmul(torch.diag_embed(sigma_corrected), Vh))  # (B,3,3)

        return F_corrected

class ElasticityModel(nn.Module):

    def __init__(
        self,
        youngs_modulus_log: float = 11.7,
        poissons_ratio_unconstrained: float = 5.5,
    ):
        """
        Corotated Elasticity model with trainable physical parameters.

        Args:
            youngs_modulus_log (float): log Young's modulus.
            poissons_ratio_unconstrained (float): unconstrained scalar for Poisson's ratio.
        """
        super().__init__()
        self.youngs_modulus_log = nn.Parameter(torch.tensor(youngs_modulus_log))  # scalar
        self.poissons_ratio_unconstrained = nn.Parameter(torch.tensor(poissons_ratio_unconstrained))  # scalar

    def forward(self, F: torch.Tensor) -> torch.Tensor:
        """
        Compute Kirchhoff stress tensor from deformation gradient tensor.

        Args:
            F (torch.Tensor): deformation gradient tensor (B, 3, 3).

        Returns:
            kirchhoff_stress (torch.Tensor): Kirchhoff stress tensor (B, 3, 3).
        """
        youngs_modulus = self.youngs_modulus_log.exp()  # scalar
        poissons_ratio = torch.sigmoid(self.poissons_ratio_unconstrained) * 0.49  # scalar in (0, 0.49)

        mu = youngs_modulus / (2.0 * (1.0 + poissons_ratio))  
        la = youngs_modulus * poissons_ratio / ((1.0 + poissons_ratio) * (1.0 - 2.0 * poissons_ratio))  

        U, sigma, Vh = torch.linalg.svd(F, full_matrices=False)  # (B,3,3), (B,3), (B,3,3)

        # Clamp singular values for numerical stability
        sigma_clamped = torch.clamp_min(sigma, 1e-5)  # (B,3)

        # Rotation matrix R = U V^T
        R = torch.matmul(U, Vh)  # (B,3,3)

        Ft = F.transpose(1, 2)  # (B,3,3)

        # Corotated stress: 2 * mu * (F - R) * F^T
        corotated_stress = 2.0 * mu * torch.matmul(F - R, Ft)  # (B,3,3)

        # Compute determinant J = product of singular values
        J = torch.prod(sigma_clamped, dim=1)  # (B,)
        J = J.view(-1, 1, 1)  # (B,1,1)

        # Identity tensor I
        I = torch.eye(3, dtype=F.dtype, device=F.device).unsqueeze(0)  # (1,3,3)

        volume_stress = la * J * (J - 1).view(-1, 1, 1) * I  # (B,3,3)

        # First Piola-Kirchhoff stress P
        P = corotated_stress + volume_stress  # (B,3,3)

        kirchhoff_stress = torch.matmul(P, Ft)  # (B,3,3)

        return kirchhoff_stress
\end{lstlisting}

\subsection{JellyDuck}
In the JellyDuck scenario, the constitutive law inferred by our method is presented.
\begin{lstlisting}[style=pythonstyle]
import torch
import torch.nn as nn

class PlasticityModel(nn.Module):

    def __init__(self, yield_stress: float = 0.1, hardening: float = 0.0):
        """
        Define trainable continuous physical parameters for differentiable optimization.
        Initialize yield stress and isotropic hardening parameters.

        Args:
            yield_stress (float): yield stress threshold for plastic correction.
            hardening (float): isotropic hardening parameter.
        """
        super().__init__()
        self.yield_stress = nn.Parameter(torch.tensor(yield_stress))  # scalar parameter
        self.hardening = nn.Parameter(torch.tensor(hardening))        # scalar parameter

    def forward(self, F: torch.Tensor) -> torch.Tensor:
        """
        Compute corrected deformation gradient using von Mises plasticity return mapping.

        Args:
            F (torch.Tensor): deformation gradient tensor (B, 3, 3).

        Returns:
            F_corrected (torch.Tensor): corrected deformation gradient tensor (B, 3, 3).
        """
        B = F.shape[0]

        # SVD of deformation gradient: F = U * diag(sigma) * Vh
        U, sigma, Vh = torch.linalg.svd(F)  # U,Vh: (B,3,3), sigma: (B,3)

        # Clamp singular values to avoid log(0)
        sigma_clamped = torch.clamp_min(sigma, 1e-5)  # (B, 3)

        # Compute logarithmic strain
        epsilon = torch.log(sigma_clamped)  # (B, 3)

        # Deviatoric strain: subtract mean (volumetric) strain
        epsilon_mean = epsilon.mean(dim=1, keepdim=True)  # (B, 1)
        epsilon_dev = epsilon - epsilon_mean  # (B, 3)

        # Norm of deviatoric strain
        epsilon_dev_norm = torch.norm(epsilon_dev, dim=1, keepdim=True)  # (B, 1)

        # Effective yield threshold with hardening, clamped to positive
        yield_threshold = torch.clamp_min(self.yield_stress + self.hardening, 1e-8)  # scalar

        # Plastic correction factor (return mapping)
        gamma = torch.clamp_min(epsilon_dev_norm - yield_threshold, 0.0) / (epsilon_dev_norm + 1e-12)  # (B,1)

        # Correct deviatoric strain
        epsilon_dev_corrected = epsilon_dev * (1 - gamma)  # (B, 3)

        # Reconstruct corrected logarithmic strain
        epsilon_corrected = epsilon_dev_corrected + epsilon_mean  # (B, 3)

        # Exponentiate to get corrected singular values
        sigma_corrected = torch.exp(epsilon_corrected)  # (B, 3)

        # Recompose corrected deformation gradient
        F_corrected = torch.matmul(U, torch.matmul(torch.diag_embed(sigma_corrected), Vh))  # (B, 3, 3)

        return F_corrected

class ElasticityModel(nn.Module):

    def __init__(self, youngs_modulus_log: float = 11.49, poissons_ratio_sigmoid: float = 1.00):
        """
        Define trainable continuous physical parameters for differentiable optimization.
        Initialize with previous best values.

        Args:
            youngs_modulus_log (float): log of Young's modulus.
            poissons_ratio_sigmoid (float): Poisson's ratio before sigmoid transformation.
        """
        super().__init__()
        self.youngs_modulus_log = nn.Parameter(torch.tensor(youngs_modulus_log))  # scalar
        self.poissons_ratio_sigmoid = nn.Parameter(torch.tensor(poissons_ratio_sigmoid))  # scalar

    def forward(self, F: torch.Tensor) -> torch.Tensor:
        """
        Compute Kirchhoff stress tensor using Corotated elasticity model.

        Args:
            F (torch.Tensor): deformation gradient tensor (B, 3, 3).

        Returns:
            kirchhoff_stress (torch.Tensor): Kirchhoff stress tensor (B, 3, 3).
        """
        B = F.size(0)

        # Recover physical parameters
        youngs_modulus = self.youngs_modulus_log.exp()  # scalar positive
        poissons_ratio = self.poissons_ratio_sigmoid.sigmoid() * 0.49  # scalar in [0, 0.49]

        mu = youngs_modulus / (2 * (1 + poissons_ratio))  # (scalar)
        la = youngs_modulus * poissons_ratio / ((1 + poissons_ratio) * (1 - 2 * poissons_ratio))  # (scalar)

        # SVD of F
        U, sigma, Vh = torch.linalg.svd(F)  # (B,3,3), (B,3), (B,3,3)
        sigma = torch.clamp_min(sigma, 1e-5)  # avoid zero singular values

        # Rotation matrix R = U * Vh
        R = torch.matmul(U, Vh)  # (B, 3, 3)

        # Determinant J = product of singular values
        J = torch.prod(sigma, dim=1).view(-1, 1, 1)  # (B, 1, 1)

        # Identity matrix I
        I = torch.eye(3, dtype=F.dtype, device=F.device).unsqueeze(0).expand(B, -1, -1)  # (B, 3, 3)

        # Corotated first Piola-Kirchhoff stress: P_corot = 2 * mu * (F - R)
        mu_expanded = mu.view(-1, 1, 1)  # (B, 1, 1)
        P_corot = 2 * mu_expanded * (F - R)  # (B, 3, 3)

        # Volume part: P_vol = la * J * (J - 1) * J * F^{-T}
        F_inv = torch.linalg.inv(F)  # (B, 3, 3)
        F_inv_T = F_inv.transpose(1, 2)  # (B, 3, 3)
        volume_factor = la.view(-1, 1, 1) * J * (J - 1).view(-1, 1, 1)  # (B, 1, 1)
        P_vol = volume_factor * J * F_inv_T  # (B, 3, 3)

        # Total first Piola-Kirchhoff stress tensor
        P = P_corot + P_vol  # (B, 3, 3)

        # Kirchhoff stress tensor tau = P @ F^T
        Ft = F.transpose(1, 2)  # (B, 3, 3)
        kirchhoff_stress = torch.matmul(P, Ft)  # (B, 3, 3)

        return kirchhoff_stress
\end{lstlisting}

\subsection{RubberPawn}
In the RubberPawn scenario, the constitutive law inferred by our method is presented.
\begin{lstlisting}[style=pythonstyle]
import torch
import torch.nn as nn

class PlasticityModel(nn.Module):

    def __init__(self, yield_stress: float = 0.22, mu_log: float = 4.0):
        """
        Define trainable continuous physical parameters for differentiable optimization.
        Initialize yield_stress and plastic shear modulus (mu) in log space.

        Args:
            yield_stress (float): yield stress controlling plastic threshold.
            mu_log (float): log shear modulus for plastic correction.
        """
        super().__init__()
        self.yield_stress = nn.Parameter(torch.tensor(yield_stress))  # scalar
        self.mu_log = nn.Parameter(torch.tensor(mu_log))  # scalar

    def forward(self, F: torch.Tensor) -> torch.Tensor:
        """
        Compute corrected deformation gradient from deformation gradient tensor via logarithmic spectral plasticity.

        Args:
            F (torch.Tensor): deformation gradient tensor (B, 3, 3).

        Returns:
            F_corrected (torch.Tensor): corrected deformation gradient tensor (B, 3, 3).
        """
        B = F.shape[0]

        mu = self.mu_log.exp()  # scalar

        # SVD decomposition
        U, sigma, Vh = torch.linalg.svd(F)  # U: (B,3,3), sigma: (B,3), Vh: (B,3,3)

        # Clamp singular values
        sigma = torch.clamp_min(sigma, 1e-6)  # (B,3)

        # Logarithmic principal stretches
        epsilon = torch.log(sigma)  # (B,3)

        # Compute volumetric mean of epsilon
        epsilon_mean = epsilon.mean(dim=1, keepdim=True)  # (B,1)

        # Deviatoric log strain
        epsilon_bar = epsilon - epsilon_mean  # (B,3)

        # Norm of deviatoric strain
        epsilon_bar_norm = torch.linalg.norm(epsilon_bar, dim=1, keepdim=True)  # (B,1)

        # Plastic multiplier
        delta_gamma = epsilon_bar_norm - self.yield_stress / (2 * mu)  # (B,1)

        # Clamp to non-negative
        delta_gamma_clamped = torch.clamp_min(delta_gamma, 0.0)  # (B,1)

        # Avoid division by zero
        denom = epsilon_bar_norm.clamp_min(1e-8)  # (B,1)

        # Compute correction scale factor
        scale = 1.0 - delta_gamma_clamped / denom  # (B,1)

        # No correction if yield condition not surpassed
        scale = torch.where(delta_gamma > 0, scale, torch.ones_like(scale))  # (B,1)

        # Apply correction
        epsilon_bar_corrected = epsilon_bar * scale  # (B,3)

        # Recompose corrected log strain
        epsilon_corrected = epsilon_bar_corrected + epsilon_mean  # (B,3)

        # Inverse log to get corrected singular values
        sigma_corrected = torch.exp(epsilon_corrected)  # (B,3)

        # Reconstructed corrected deformation gradient
        F_corrected = U @ torch.diag_embed(sigma_corrected) @ Vh  # (B,3,3)

        return F_corrected

class ElasticityModel(nn.Module):

    def __init__(self, youngs_modulus_log: float = 12.9, poissons_ratio_sigmoid: float = 0.0):
        """
        Define trainable continuous physical parameters for differentiable optimization.
        Initialize parameters from best prior estimates.

        Args:
            youngs_modulus_log (float): log of Young's modulus.
            poissons_ratio_sigmoid (float): raw Poisson's ratio parameter before sigmoid scaling.
        """
        super().__init__()
        self.youngs_modulus_log = nn.Parameter(torch.tensor(youngs_modulus_log))  # scalar
        self.poissons_ratio_sigmoid = nn.Parameter(torch.tensor(poissons_ratio_sigmoid))  # scalar

    def forward(self, F: torch.Tensor) -> torch.Tensor:
        """
        Compute Kirchhoff stress from corrected deformation gradient tensor using StVK elasticity.

        Args:
            F (torch.Tensor): deformation gradient tensor (B, 3, 3).

        Returns:
            kirchhoff_stress (torch.Tensor): Kirchhoff stress tensor (B, 3, 3).
        """
        B = F.shape[0]

        # Physical parameters
        youngs_modulus = self.youngs_modulus_log.exp()  # scalar

        # Sigmoid mapping to (0, 0.499) for Poisson's ratio
        poissons_ratio = torch.sigmoid(self.poissons_ratio_sigmoid) * 0.499  # scalar

        mu = youngs_modulus / (2.0 * (1.0 + poissons_ratio))  # scalar
        la = youngs_modulus * poissons_ratio / ((1.0 + poissons_ratio) * (1.0 - 2.0 * poissons_ratio))  # scalar

        I = torch.eye(3, dtype=F.dtype, device=F.device).unsqueeze(0)  # (1, 3, 3)

        Ft = F.transpose(1, 2)  # (B, 3, 3)

        # Right Cauchy-Green tensor
        C = torch.matmul(Ft, F)  # (B, 3, 3)

        # Green-Lagrange strain tensor
        E = 0.5 * (C - I)  # (B, 3, 3)

        # Trace of strain tensor
        trE = E.diagonal(dim1=1, dim2=2).sum(dim=1).view(B, 1, 1)  # (B, 1, 1)

        # Second Piola-Kirchhoff stress tensor
        S = 2.0 * mu * E + la * trE * I  # (B, 3, 3)

        # First Piola-Kirchhoff stress tensor
        P = torch.matmul(F, S)  # (B, 3, 3)

        # Kirchhoff stress tensor: tau = P * F^T
        kirchhoff_stress = torch.matmul(P, Ft)  # (B, 3, 3)

        return kirchhoff_stress
\end{lstlisting}

\subsection{SandFish}
In the SandFish scenario, the constitutive law inferred by our method is presented.
\begin{lstlisting}[style=pythonstyle]
import torch
import torch.nn as nn

class PlasticityModel(nn.Module):

    def __init__(self, yield_stress: float = 0.07):
        """
        Define trainable plastic yield stress parameter with enforced numerical stability.

        Args:
            yield_stress (float): yield stress controlling deviatoric plastic flow magnitude.
        """
        super().__init__()
        self.yield_stress = nn.Parameter(torch.tensor(yield_stress))

    def forward(self, F: torch.Tensor) -> torch.Tensor:
        """
        Compute plasticity-corrected deformation gradient by shrinking deviatoric logarithmic strain.

        Args:
            F (torch.Tensor): deformation gradient tensor (B, 3, 3).

        Returns:
            F_corrected (torch.Tensor): corrected deformation gradient tensor (B, 3, 3).
        """
        # SVD decomposition
        U, sigma, Vh = torch.linalg.svd(F)                            # (B, 3, 3), (B, 3), (B, 3, 3)

        # Clamp singular values for stability
        sigma_clamped = torch.clamp_min(sigma, 1e-6)                  # (B, 3)

        # Compute logarithmic principal strain
        epsilon = torch.log(sigma_clamped)                            # (B, 3)

        # Volumetric part (mean)
        epsilon_mean = epsilon.mean(dim=1, keepdim=True)              # (B, 1)

        # Deviatoric strain
        epsilon_dev = epsilon - epsilon_mean                          # (B, 3)

        # Norm of deviatoric strain
        epsilon_dev_norm = torch.linalg.norm(epsilon_dev, dim=1, keepdim=True)  # (B, 1)

        # Enforce minimum yield stress to avoid numerical instability
        yield_stress = torch.clamp_min(self.yield_stress, 0.05)       # scalar

        # Clamp norm for division
        epsilon_dev_norm_safe = torch.clamp_min(epsilon_dev_norm, 1e-12)        # (B, 1)

        # Compute plastic correction magnitude delta_gamma
        delta_gamma = epsilon_dev_norm - yield_stress                  # (B, 1)
        delta_gamma_clamped = torch.clamp_min(delta_gamma, 0.0)        # (B, 1)

        # Scaling factor for deviatoric strain correction
        scale = 1.0 - delta_gamma_clamped / epsilon_dev_norm_safe       # (B, 1)
        scale = torch.clamp_min(scale, 0.0)                            # (B, 1)

        # Apply plastic correction to deviatoric strain
        epsilon_dev_corrected = epsilon_dev * scale                     # (B, 3)

        # Recombine volumetric and deviatoric parts
        epsilon_corrected = epsilon_mean + epsilon_dev_corrected       # (B, 3)

        # Calculate corrected singular values
        sigma_corrected = torch.exp(epsilon_corrected)                  # (B, 3)

        # Reconstruct corrected deformation gradient
        F_corrected = U @ torch.diag_embed(sigma_corrected) @ Vh       # (B, 3, 3)

        return F_corrected

class ElasticityModel(nn.Module):

    def __init__(self, youngs_modulus_log: float = 9.55, poissons_ratio_sigmoid: float = 2.50):
        """
        Define trainable Young's modulus and Poisson's ratio with physically realistic bounds.

        Args:
            youngs_modulus_log (float): logarithm of Young's modulus.
            poissons_ratio_sigmoid (float): raw parameter to be passed through sigmoid for Poisson's ratio.
        """
        super().__init__()
        self.youngs_modulus_log = nn.Parameter(torch.tensor(youngs_modulus_log))
        self.poissons_ratio_sigmoid = nn.Parameter(torch.tensor(poissons_ratio_sigmoid))

    def forward(self, F: torch.Tensor) -> torch.Tensor:
        """
        Compute Kirchhoff stress tensor from deformation gradient with corotated elasticity.

        Args:
            F (torch.Tensor): deformation gradient tensor (B, 3, 3).

        Returns:
            kirchhoff_stress (torch.Tensor): Kirchhoff stress tensor (B, 3, 3).
        """
        B = F.shape[0]

        # Recover material parameters
        E = self.youngs_modulus_log.exp()                              # scalar
        nu_raw = self.poissons_ratio_sigmoid.sigmoid()                 # (0,1)
        nu = nu_raw * 0.45                                              # scale to max 0.45 Poisson ratio (~stable and compressible)

        mu = E / (2.0 * (1.0 + nu))                                    # scalar
        lam = E * nu / ((1.0 + nu) * (1.0 - 2.0 * nu))                 # scalar

        # Compute SVD
        U, sigma, Vh = torch.linalg.svd(F)                            # (B, 3, 3), (B, 3), (B, 3, 3)

        # Clamp singular values to prevent numerical issues
        sigma_clamped = torch.clamp_min(sigma, 1e-6)                  # (B, 3)

        # Compute rotation part R
        R = U @ Vh                                                    # (B, 3, 3)

        # Expand mu for broadcasting
        if mu.dim() > 0:
            mu_expanded = mu.view(-1, 1, 1)                         # (B, 1, 1)
        else:
            mu_expanded = mu                                          # scalar

        # Corotated stress part: 2 * mu * (F - R)
        corotated_stress = 2.0 * mu_expanded * (F - R)               # (B, 3, 3)

        # Compute determinant J and clamp for stability
        J = torch.linalg.det(F)                                       # (B,)
        J_clamped = torch.clamp_min(J, 1e-8)                         # (B,)

        # Identity tensor I (1, 3, 3)
        I = torch.eye(3, dtype=F.dtype, device=F.device).unsqueeze(0) # (1, 3, 3)

        # Expand and reshape parameters for broadcasting
        if lam.dim() > 0:
            lam_expanded = lam.view(-1, 1, 1)                        # (B, 1, 1)
        else:
            lam_expanded = lam                                         # scalar

        J_expanded = J_clamped.view(-1, 1, 1)                         # (B, 1, 1)
        J_minus_1_expanded = (J_clamped - 1.0).view(-1, 1, 1)         # (B, 1, 1)

        # Volumetric stress: lambda * J * (J - 1) * I
        volumetric_stress = lam_expanded * J_expanded * J_minus_1_expanded * I  # (B, 3, 3)

        # First Piola-Kirchhoff stress
        P = corotated_stress + volumetric_stress                      # (B, 3, 3)

        # Transpose of deformation gradient
        Ft = F.transpose(1, 2)                                        # (B, 3, 3)

        # Kirchhoff stress tensor: tau = P @ F^T
        kirchhoff_stress = P @ Ft                                     # (B, 3, 3)

        return kirchhoff_stress
\end{lstlisting}

\subsection{Bun}
In the Bun scenario, the constitutive law inferred by our method is presented.
\begin{lstlisting}[style=pythonstyle]
import torch
import torch.nn as nn

class PlasticityModel(nn.Module):
    def __init__(self, yield_stress: float = 0.30):
        """
        Trainable continuous yield stress parameter for von Mises plasticity correction.

        Args:
            yield_stress (float): yield stress threshold for plastic correction.
        """
        super().__init__()
        self.yield_stress = nn.Parameter(torch.tensor(yield_stress))

    def forward(self, F: torch.Tensor) -> torch.Tensor:
        """
        Compute corrected deformation gradient from input deformation gradient tensor.

        Args:
            F (torch.Tensor): deformation gradient tensor (B, 3, 3).

        Returns:
            F_corrected (torch.Tensor): corrected deformation gradient tensor (B, 3, 3).
        """
        # Compute SVD of F: U, sigma, Vh
        U, sigma, Vh = torch.linalg.svd(F)  # U: (B,3,3), sigma: (B,3), Vh: (B,3,3)

        # Clamp singular values to avoid log(0)
        sigma_clamped = torch.clamp_min(sigma, 1e-6)  # (B,3)

        # Compute logarithm of singular values (principal logarithmic strains)
        epsilon = torch.log(sigma_clamped)  # (B,3)

        # Compute volumetric mean strain
        epsilon_mean = epsilon.mean(dim=1, keepdim=True)  # (B,1)

        # Deviatoric strain (deviation from mean)
        epsilon_dev = epsilon - epsilon_mean  # (B,3)

        # Norm of deviatoric strain, clamp to avoid numerical issues
        epsilon_dev_norm = torch.norm(epsilon_dev, dim=1, keepdim=True).clamp_min(1e-12)  # (B,1)

        # Compute plastic multiplier (excess over yield stress)
        delta_gamma = epsilon_dev_norm - self.yield_stress  # (B,1)

        # Apply plastic correction only if exceeding yield stress
        delta_gamma_clamped = torch.clamp_min(delta_gamma, 0.0)  # (B,1)

        # Calculate shrink factor for deviatoric strains
        shrink_factor = 1.0 - delta_gamma_clamped / epsilon_dev_norm  # (B,1)

        # Correct deviatoric strain by projecting onto yield surface
        epsilon_dev_corrected = epsilon_dev * shrink_factor  # (B,3)

        # Reassemble corrected total logarithmic strains
        epsilon_corrected = epsilon_mean + epsilon_dev_corrected  # (B,3)

        # Exponentiate to get corrected singular values
        sigma_corrected = torch.exp(epsilon_corrected)  # (B,3)

        # Reconstruct corrected deformation gradient: F_corrected = U * diag(sigma_corrected) * Vh
        F_corrected = U @ torch.diag_embed(sigma_corrected) @ Vh  # (B,3,3)

        return F_corrected

class ElasticityModel(nn.Module):
    def __init__(self, youngs_modulus_log: float = 9.82, poissons_ratio_sigmoid: float = 4.07):
        """
        Trainable continuous parameters for Neo-Hookean elasticity.

        Args:
            youngs_modulus_log (float): log of Young's modulus.
            poissons_ratio_sigmoid (float): Poisson's ratio parameter before sigmoid scaling.
        """
        super().__init__()
        self.youngs_modulus_log = nn.Parameter(torch.tensor(youngs_modulus_log))
        self.poissons_ratio_sigmoid = nn.Parameter(torch.tensor(poissons_ratio_sigmoid))

    def forward(self, F: torch.Tensor) -> torch.Tensor:
        """
        Compute Kirchhoff stress tensor from deformation gradient tensor using Neo-Hookean elasticity.

        Args:
            F (torch.Tensor): deformation gradient tensor (B, 3, 3).

        Returns:
            kirchhoff_stress (torch.Tensor): Kirchhoff stress tensor (B, 3, 3).
        """
        B = F.size(0)  # batch size

        # Compute Young's modulus E and Poisson's ratio nu
        E = self.youngs_modulus_log.exp()  # scalar
        nu = self.poissons_ratio_sigmoid.sigmoid() * 0.49  # scalar in (0,0.49)

        mu = E / (2 * (1 + nu))  # scalar
        lam = E * nu / ((1 + nu) * (1 - 2 * nu))  # scalar

        # Identity tensor I (B,3,3)
        I = torch.eye(3, dtype=F.dtype, device=F.device).unsqueeze(0).expand(B, -1, -1)  # (B,3,3)

        # Compute determinant J of F (B,)
        J = torch.linalg.det(F).clamp_min(1e-12).view(-1, 1, 1)  # (B,1,1)
        logJ = torch.log(J)  # (B,1,1)

        # Compute inverse transpose of F (B,3,3)
        F_inv = torch.inverse(F)  # (B,3,3)
        F_inv_T = F_inv.transpose(1, 2)  # (B,3,3)

        # Compute first Piola-Kirchhoff stress tensor P = mu*(F - F_inv_T) + lam*logJ*F_inv_T
        P = mu * (F - F_inv_T) + lam * logJ * F_inv_T  # (B,3,3)

        # Compute Kirchhoff stress tau = P * F^T
        Ft = F.transpose(1, 2)  # (B,3,3)
        kirchhoff_stress = torch.matmul(P, Ft)  # (B,3,3)

        return kirchhoff_stress
\end{lstlisting}

\subsection{Burger}
In the Burger scenario, the constitutive law inferred by our method is presented.
\begin{lstlisting}[style=pythonstyle]
import torch
import torch.nn as nn

class PlasticityModel(nn.Module):

    def __init__(self):
        """
        Identity plasticity: no correction to deformation gradient.
        """
        super().__init__()

    def forward(self, F: torch.Tensor) -> torch.Tensor:
        """
        Args:
            F (torch.Tensor): deformation gradient tensor (B, 3, 3).

        Returns:
            F_corrected (torch.Tensor): corrected deformation gradient tensor (B, 3, 3).
        """
        # No plastic correction
        return F  # (B, 3, 3)

class ElasticityModel(nn.Module):

    def __init__(self,
                 youngs_modulus_log: float = 8.37,
                 poissons_ratio: float = 0.49):
        """
        Corotated elasticity with trainable parameters.

        Args:
            youngs_modulus_log (float): log of Young's modulus.
            poissons_ratio (float): Poisson's ratio (clamped [0,0.49]).
        """
        super().__init__()
        self.youngs_modulus_log = nn.Parameter(torch.tensor(youngs_modulus_log))
        self.poissons_ratio = nn.Parameter(torch.tensor(poissons_ratio))

    def forward(self, F: torch.Tensor) -> torch.Tensor:
        """
        Compute Kirchhoff stress tensor from deformation gradient tensor.

        Args:
            F (torch.Tensor): deformation gradient tensor (B, 3, 3).

        Returns:
            kirchhoff_stress (torch.Tensor): Kirchhoff stress tensor (B, 3, 3).
        """
        B = F.shape[0]

        # Physical parameters
        E = self.youngs_modulus_log.exp()  # scalar
        nu = torch.clamp(self.poissons_ratio, 0.0, 0.49)  # scalar

        mu = E / (2.0 * (1.0 + nu))  # scalar
        la = E * nu / ((1.0 + nu) * (1.0 - 2.0 * nu))  # scalar

        # SVD of F: U, Sigma, Vh such that F = U @ diag(Sigma) @ Vh
        U, sigma, Vh = torch.linalg.svd(F)  # U: (B,3,3), sigma: (B,3), Vh: (B,3,3)
        sigma = torch.clamp_min(sigma, 1e-5)  # (B,3) ensure positivity

        # Rotation R = U @ Vh
        R = torch.matmul(U, Vh)  # (B,3,3)

        # Corotated stress part: tau_c = 2*mu*(F - R) @ F^T
        Ft = F.transpose(1, 2)  # (B,3,3)
        tau_c = 2.0 * mu * torch.matmul(F - R, Ft)  # (B,3,3)

        # Volumetric part: tau_v = lambda * J * (J - 1) * I
        J = torch.prod(sigma, dim=1).view(B, 1, 1)  # (B,1,1)
        I = torch.eye(3, dtype=F.dtype, device=F.device).unsqueeze(0).expand(B, -1, -1)  # (B,3,3)
        tau_v = la * J * (J - 1) * I  # (B,3,3)

        # Kirchhoff stress
        kirchhoff_stress = tau_c + tau_v  # (B,3,3)

        return kirchhoff_stress
\end{lstlisting}

\end{document}

%% file: math_commands.tex
\usepackage{amsmath,amsfonts,bm}

\def\eqref#1{equation~\ref{#1}}
\def\1{\bm{1}}

\DeclareMathAlphabet{\mathsfit}{\encodingdefault}{\sfdefault}{m}{sl}
\SetMathAlphabet{\mathsfit}{bold}{\encodingdefault}{\sfdefault}{bx}{n}